%% file: asap21.tex
\documentclass[conference]{IEEEtran}
\IEEEoverridecommandlockouts
\usepackage{cite}
\usepackage{amsmath,amssymb,amsfonts}
\usepackage{algorithmic}
\usepackage{graphicx}
\usepackage{textcomp}
\usepackage{xcolor}
\def\BibTeX{{\rm B\kern-.05em{\sc i\kern-.025em b}\kern-.08em
    T\kern-.1667em\lower.7ex\hbox{E}\kern-.125emX}}
    
\usepackage{multirow}
\usepackage{gensymb}
\usepackage{threeparttable}
\usepackage[caption=false]{subfig}
\newcommand{\ineq}[1]{\footnotesize$#1$\normalsize}{}
\newcommand{\sm}{\text{{SpiNeMap}}}{}

\renewcommand{\baselinestretch}{0.99}
    
\begin{document}
\bstctlcite{IEEEexample:BSTcontrol}

\title{Endurance Management in Neuromorphic Computing via Hardware-Software Co-design}
\title{Improving Endurance of Neuromorphic Hardware via Hardware-Software Co-design}
\title{Improving Read Endurance of Neuromorphic Systems via Intelligent Synapse Mapping}
\title{Improving Inference Lifetime of Neuromorphic Systems via Intelligent Synapse Mapping}

\author{\IEEEauthorblockN{Shihao Song, Twisha Titirsha, and Anup Das}
\IEEEauthorblockA{{Electrical and Computer Engineering, Drexel University,} 
Philadelphia, PA \\
\{shihao.song,tt624,anup.das\}@drexel.edu}
}

\maketitle

\begin{abstract}
\input{sections/abstract}
\end{abstract}

\begin{IEEEkeywords}
Neuromorphic Computing, Non-Volatile Memory (NVM), Endurance, RRAM, Spiking Neural Network (SNN)
\end{IEEEkeywords}

\section{Introduction}\label{sec:introduction}
\input{sections/introduction}

\section{Background}\label{sec:background}
\input{sections/background}

\section{Variation of Read Endurance}\label{sec:endurance_variation}
\input{sections/endurance_variation}

\section{Problem Formulation}\label{sec:problem_formulation}
\input{sections/problem_formulation}

\section{Results and Discussion}\label{sec:results}
\input{sections/results}

\section{Conclusions}\label{sec:conclusions}

\input{sections/conclusions}

\section*{Acknowledgment}
This work is supported by the National Science Foundation Faculty Early Career Development Award CCF-1942697.

\bibliographystyle{IEEEtran}
\IEEEtriggeratref{11}
\bibliography{commands,disco,external}

\end{document}

%% file: sections/abstract.tex
Non-Volatile Memories (NVMs) such as Resistive RAM (RRAM) are used in neuromorphic systems to implement high-density and low-power analog synaptic weights.
Unfortunately, an RRAM cell can switch its state after reading its content a certain number of times. 
Such behavior challenges the integrity and program-once-read-many-times philosophy of implementing machine learning inference on neuromorphic systems, impacting the Quality-of-Service (QoS). Elevated temperatures and frequent usage can significantly shorten the number of times an RRAM cell can be reliably read before it becomes absolutely necessary to reprogram. 
We propose an architectural solution to extend the read endurance of RRAM-based neuromorphic systems. We make two key contributions. First, we formulate the read endurance of an RRAM cell as a function of the programmed synaptic weight and its activation within a machine learning workload. 
Second, we propose an intelligent workload mapping strategy incorporating the endurance formulation to place the synapses of a machine learning model onto the RRAM cells of the hardware. The objective is to extend the inference lifetime, defined as the number of times 
the model can be used to generate output (inference) before the trained weights need to be reprogrammed on the RRAM cells of the system.
We evaluate our architectural solution with machine learning workloads on a cycle-accurate simulator of an RRAM-based neuromorphic system. Our results demonstrate a significant increase in inference lifetime with only a minimal performance impact.

%% file: sections/introduction.tex
Neuromorphic systems are integrated circuits that mimic the neuro-biological architecture of the central nervous system~\cite{mead1990neuromorphic}. 
They employ variants of integrate-and-fire (I\&F) neurons as computational units and analog weights as synaptic storage. 
I\&F neurons use spikes to encode information, where each spike is a voltage or current pulse, typically of an ms duration~\cite{maass1997networks}.
Due to its event (spike)-driven operations, a neuromorphic system consumes less power and therefore, well suited as the hardware for inference of trained machine learning models deployed in power-constrained environments such as Embedded Systems and Internet-of-Things (IoT).

Non-Volatile Memory (NVM) technologies such as Filamentary Oxide-based Resistive RAM (RRAM), Phase-Change Memory (PCM), and Spin-based Magnetic RAM (MRAM) enable  low-voltage multilevel operations, making them suitable for implementing analog synaptic weight storage in neuromorphic systems~\cite{Burr2017,mallik2017design,catthoor2018very}.
Of these emerging new memory technologies, hafnia (HfO\ineq{{}_2})-based RRAM has shown a significant promise due to its CMOS compatibility at scaled nodes, allowing the fabrication of high-density synaptic storage for neuromorphic systems. The synaptic weights are programmed on RRAM cells as conductance. An RRAM cell can be programmed to a high-resistance state (HRS) or one of the low-resistance states (LRS). 
Unfortunately, RRAM cells have limited \emph{read endurance}, i.e., an RRAM cell can switch its state after performing a certain number of reads~\cite{shim2020impact}.
To give an example, a single quasi-static read for 5000 ms or 5000 reads with 1-ms read time can lead to an abrupt change from HRS to LRS state in an RRAM cell~\cite{shim2020impact}. To put in the context of neuromorphic computing, an RRAM cell can reliably propagate 5000 1-ms spikes before it becomes absolutely necessary to reprogram the state of the cell.

\begin{figure}[h!]
	\begin{center}
		\vspace{-10pt}
		\includegraphics[width=0.99\columnwidth]{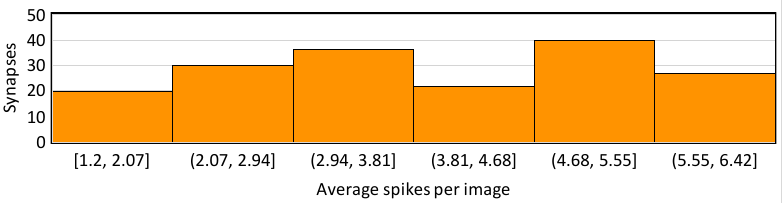}
		\vspace{-20pt}
		\caption{Spike distribution across the synapses of VGG.}
		\label{fig:vgg_distribution}
		\vspace{-15pt}
	\end{center}
\end{figure}

We now extrapolate this RRAM device behavior to the application level, describing such extrapolation with the running example of VGG, a deep learning model trained on CIFAR-10 dataset and performing inference on an RRAM-based neuromorphic system. Figure~\ref{fig:vgg_distribution} shows the histogram of average spikes per image propagating through the synapses of VGG. We collected these statistics by analyzing CIFAR-10 training and test datasets. 
We see that some synapses propagate more spikes than others when inferring an image.
These are called the \emph{critical synapses} and they decide how many images can be reliably inferred using the VGG model before 
it becomes necessary to reprogram the trained synaptic weights on the RRAM cells of the hardware.

To give an example, assume \ineq{n} to be the maximum number of spikes per image on the critical synapses of VGG (\ineq{n = 6.42} in Figure~\ref{fig:vgg_distribution}). Then, the RRAM cells need to be reprogrammed 
once every \ineq{\frac{5000}{n} \approx 778} images
to ensure correctness. The time to infer \ineq{778} images is called the \emph{inference lifetime}. 
Formally,
\begin{equation}
    \label{eq:lifetime_computation}
    \footnotesize \text{Inference Lifetime} = \frac{\text{Read Endurance}}{\text{spikes per image}}
\end{equation}

If the RRAM devices implementing VGG are not reprogrammed before the inference lifetime expires, then the accuracy of VGG can drop significantly (accuracy in the low 20\% is reported in~\cite{shim2020impact}). 

Periodic reprogramming of synaptic weights on a neuromorphic system challenges the program-once-read-many-times philosophy of machine learning inference hardware, which can impose significant system overhead. To give an example, imagine such systems are deployed at the edge nodes of an IoT infrastructure. Frequent updates of these nodes with trained weights will 1) increase communication between the edge and cloud, and 2) reduce the Quality-of-Service (QoS) due to offlining of the edge nodes every time they are reprogrammed.

We observe that inside a neuromorphic system, the RRAM cells are organized into crossbars. The parasitic IR drops in a crossbar create a difference in the voltage needed to propagate spike through the RRAM cells in the crossbar~\cite{twisha_thermal}. Such voltage differences create a variation of read endurance of the RRAM cells,
i.e., some RRAM cells are stronger than others, where the \textit{strength} of an RRAM cell is measured in terms of its read endurance, which is a function of the voltage. 
Unfortunately, if the critical synapses (those that propagate more spikes) are mapped on weaker RRAM cells (those that have low read endurance), then the inference lifetime can decrease significantly, lowering the QoS.
We propose an \textit{intelligent synapse allocation strategy}, which analyzes spikes propagating through each synapse of a machine learning model during inference and uses such information to map the model's synaptic weights to the RRAM cells considering the variation in their read endurance. The objective is to maximize the inference lifetime of the hardware. Our architectural solution is built on the following three key \textbf{contributions}.
\begin{itemize}
    \item First, we investigate the internal architecture of an RRAM-based neuromorphic system and estimate the endurance variation through detailed circuit-level simulations at different process and temperature corners.
    \item Second, we analyze a trained machine learning model and estimate the spikes propagating through its synapses.
    \item Finally, we use a Hill-Climbing approach that uses Binary Non-Linear Programming (BNLP) to map the synapses of a machine learning model to the RRAM cells such that the critical synapses are always mapped to stronger RRAM cells, thereby improving the inference lifetime.
\end{itemize}

We evaluate our architectural approach with different machine learning models on NeuroXplorer~\cite{neuroxplorer}, a cycle-accurate simulator of RRAM-based neuromorphic system. Results show a significant improvement in inference lifetime with a minimal impact on model performance.

%% file: sections/background.tex
\begin{figure}[h!]%
    \centering
    \subfloat[Tiled neuromorphic hardware.\label{fig:system_arch}]{{\includegraphics[width=4.0cm]{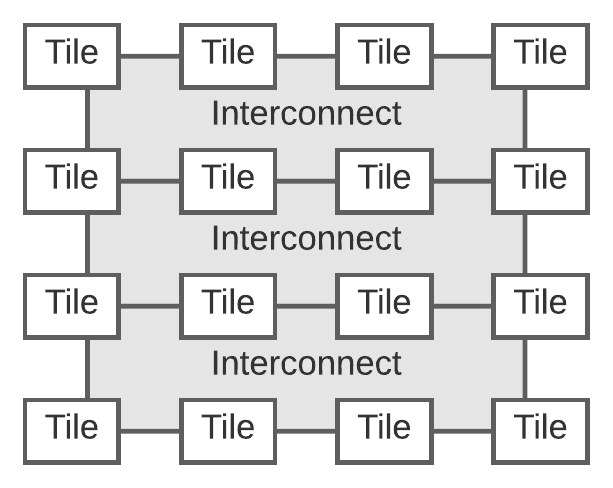} }}%
    \subfloat[Crossbar architecture.\label{fig:crossbar}]{{\includegraphics[width=3.8cm]{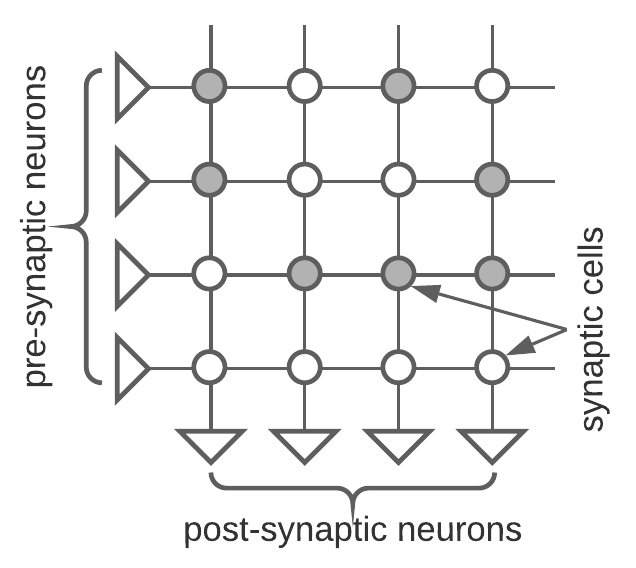} }}%
    \quad
    \subfloat[A 3-D view of a crossbar.\label{fig:3d_view}]{{\includegraphics[width=4.0cm]{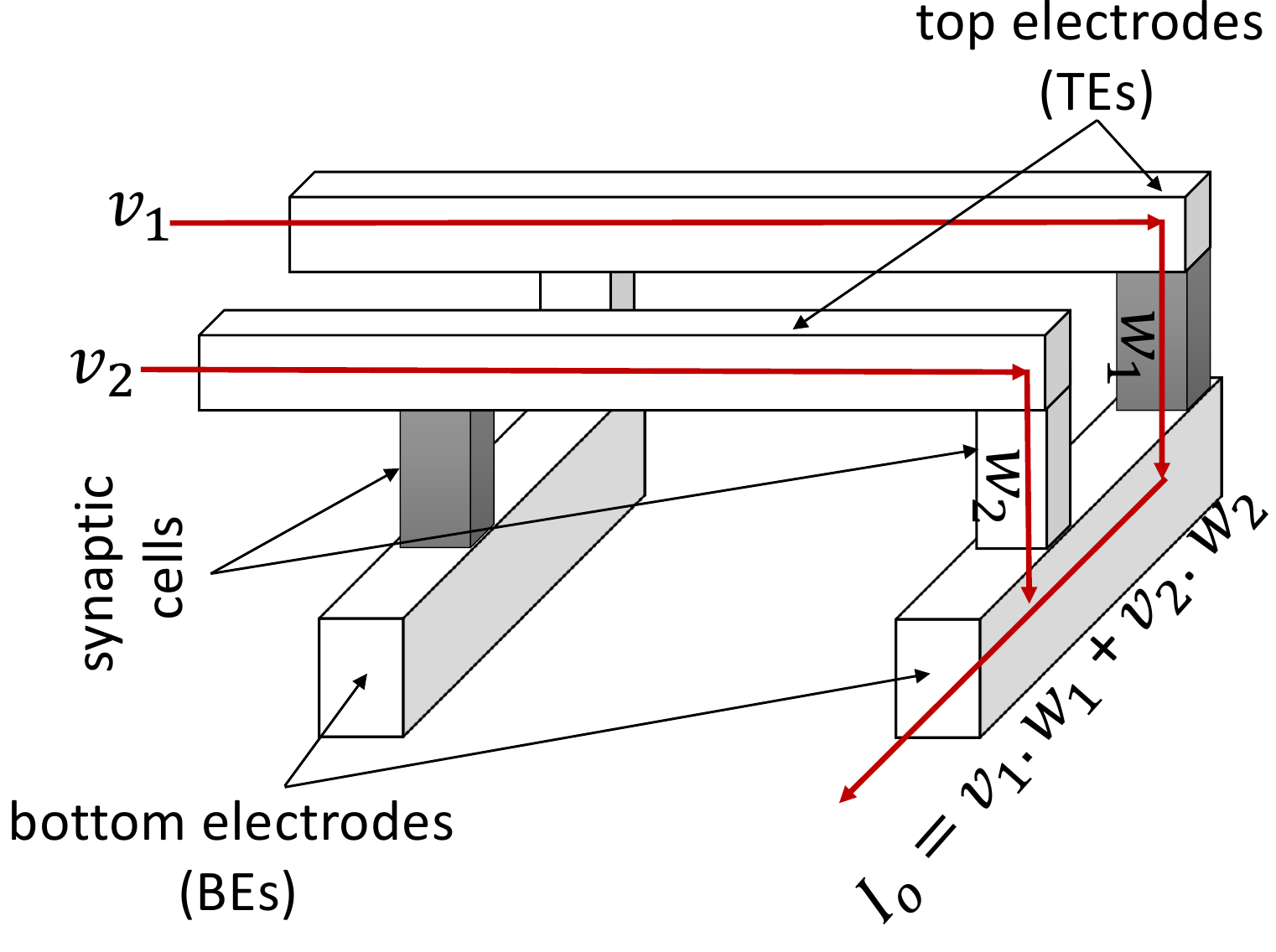} }}%
    \subfloat[Parasitics of a crossbar.\label{fig:parasitics}]{{\includegraphics[width=4.0cm]{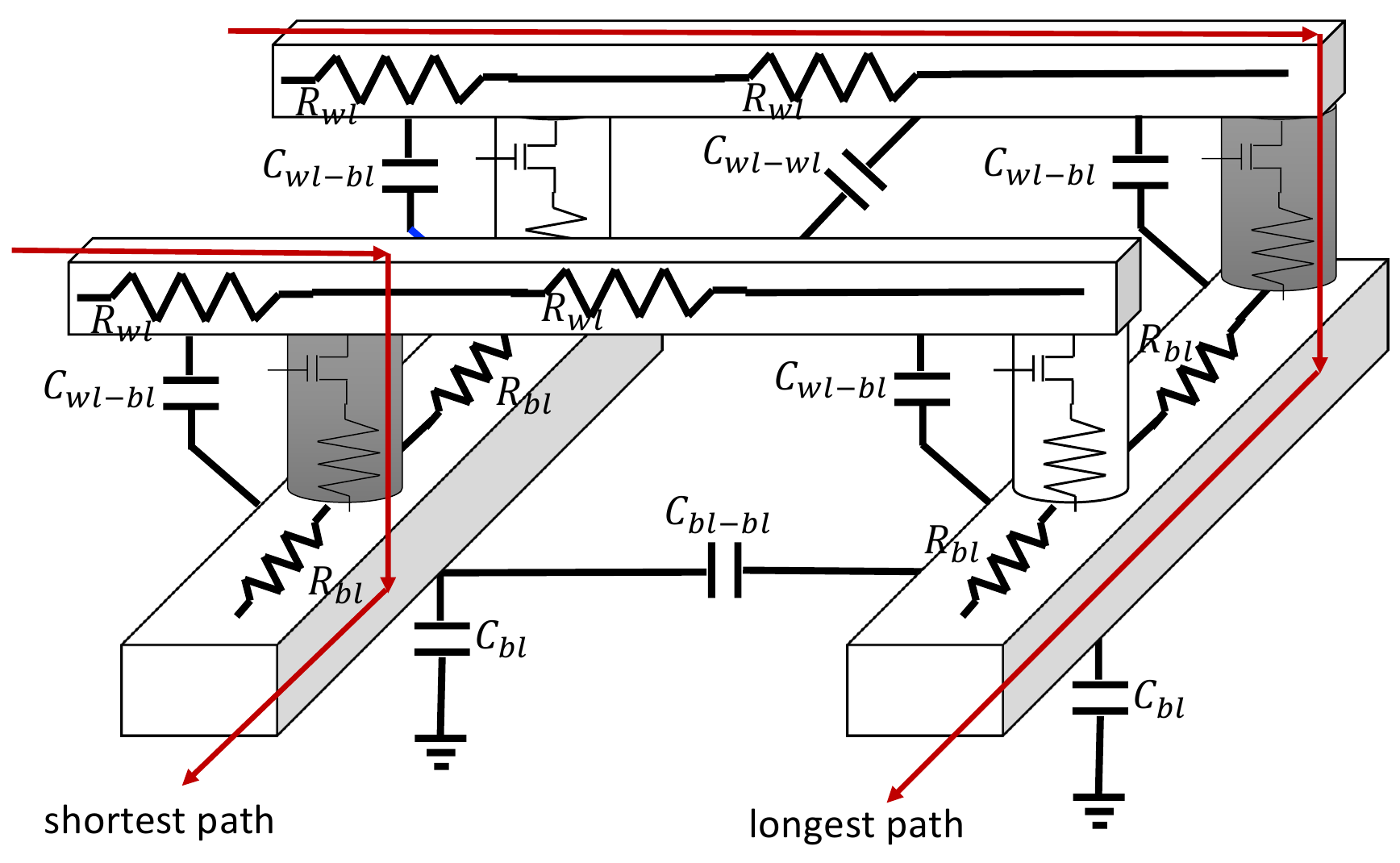} }}%
    \caption{Neuromorphic system architecture with crossbars.}%
    \label{fig:system_architecture}%
\end{figure}

A neuromorphic system is implemented as a tiled architecture (see Fig.~\ref{fig:system_arch}), where the tiles are interconnected hierarchically using a shared interconnect such as Network-on-Chip (Noc)~\cite{liu2018neu} or Segmented Bus~\cite{balaji2019exploration}. This is the representative architecture of many recent systems such as TrueNorth~\cite{truenorth}, Loihi~\cite{loihi}, and DYNAPs~\cite{dynapse}. 
In many recent systems, a tile is implemented using a crossbar, which is illustrated in Figure~\ref{fig:crossbar}. An \ineq{M}x\ineq{M} crossbar can accommodate \ineq{M} pre-synaptic neurons, mapped along the rows and \ineq{M} post-synaptic neurons, mapped along the columns. There are \ineq{{M}^2} synaptic cells, which store the weights. Figure~\ref{fig:3d_view} shows a 2x2 crossbar in three-dimension, with the top electrodes forming the rows and bottom electrodes forming the columns. A synaptic cell is placed at each intersection of top and bottom electrodes. 

\vspace{-20pt}
\vspace{-10pt}

Figure~\ref{fig:parasitics} shows the different parasitic components inside a crossbar. Such components cause variable delays on the current paths inside the crossbar. For simplicity, we have only shown the current on the shortest and the longest paths in the crossbar, where the length of a current path is measured in terms of the number of parasitic elements on the path. Therefore, spike propagation delay through synapses on longer paths is higher than on shorter paths. Although optimizing inference lifetime is our primary focus, we also evaluate the impact of our architectural solution on spike propagation delay (see Section~\ref{sec:results}). Parasitic components in a crossbar also lead to voltage variations, which impact read endurance of the synaptic cells. We analyze such impact in Section~\ref{sec:endurance_variation}.

\subsection{Machine Learning Inference on Neuromorphic Systems}\label{sec:clustering}
\begin{figure}[h!]%
    \centering
    \subfloat[VGG Convolution Neural Network (CNN).\label{fig:vgg}]{{\includegraphics[width=6.7cm]{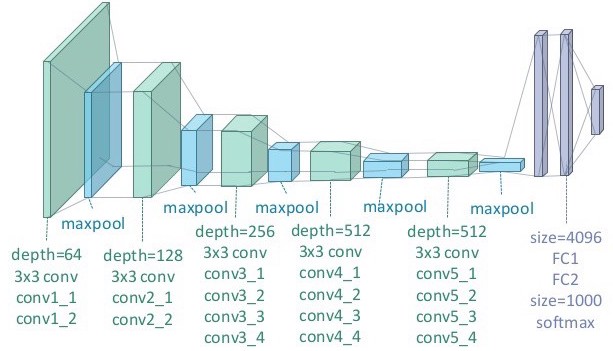} }}%
    \quad
    \subfloat[First 10 clusters of VGG (out of 95,452) clusters).\label{fig:cluster_vgg}]{{\includegraphics[width=6.7cm]{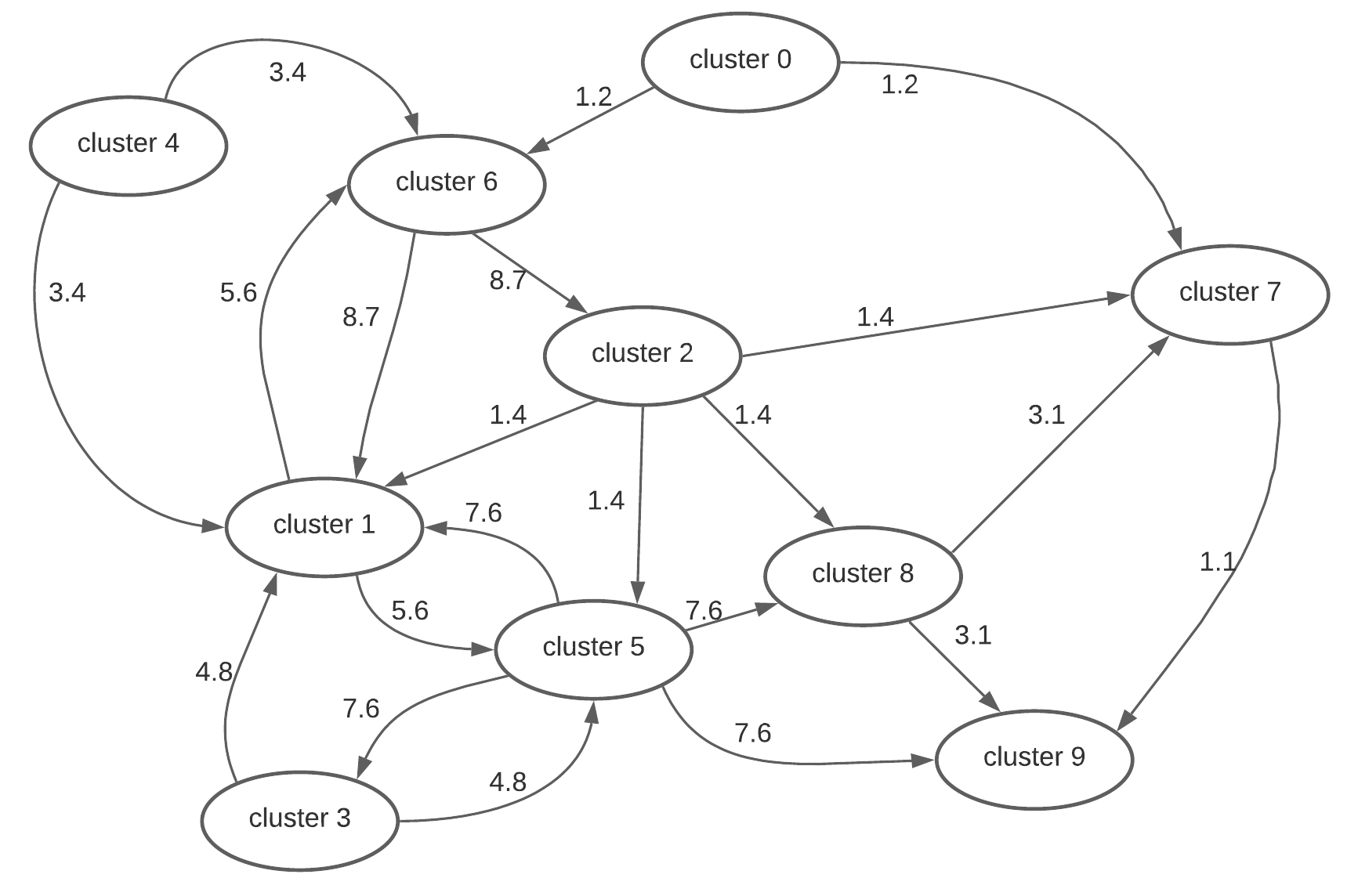} }}%
    \caption{Trained VGG model and its clusters generated using SpiNeMap~\cite{spinemap}.}%
    \label{fig:vggnet_example}%
\end{figure}

Each crossbar in a neuromorphic system can accommodate only a limited number of neurons and synapses.
To map large models, the model is first partitioned into clusters of neurons and synapses, where each cluster can fit onto a crossbar of the hardware~\cite{spinemap,esl20,dfsynthesizer,pycarl,balaji2019design,ji2016neutrams,adarsha_igsc}.
Figure~\ref{fig:vgg} shows the architecture of VGG for CIFAR-10 classification. Figure~\ref{fig:cluster_vgg} shows the first 10 clusters generated using SpiNeMap~\cite{spinemap}, a state-of-the-art approach to map machine learning inference to neuromorphic systems.
The figure illustrates the connections between these clusters, with the number on edge representing the average number of spikes communicated between the source and destination clusters when processing an image during inference.

When partitioning a machine learning model into clusters, SpiNeMap aims to minimize the inter-cluster spikes, which reduces the energy consumption on the interconnect of the hardware. There are also other optimization objectives proposed in literature. Examples include improving crossbar utilization~\cite{ji2016neutrams}, reducing crossbar usage~\cite{esl20}, reducing energy consumption~\cite{twisha_energy,psopart,twisha_thermal,das2018dataflow,balaji2019ISVLSIframework,balaji2020run}, and reducing circuit aging~\cite{reneu,vts_das,balaji2019framework,song2020case,ncrtm}. None of these approaches address mapping of the synapses of a cluster to the synaptic cells of a crossbar for the purpose of inference on neuromorphic systems. To understand why such mapping matters, we now introduce the background on filamentary oxide-based RRAM technology, which can be used to design the synaptic cells of a crossbar.

\vspace{-10pt}
\vspace{-10pt}

\subsection{Oxide-based Resistive RAM (RRAM) Technology}
\begin{figure}[h!]
	\begin{center}
		\vspace{-10pt}
		\includegraphics[width=0.6\columnwidth]{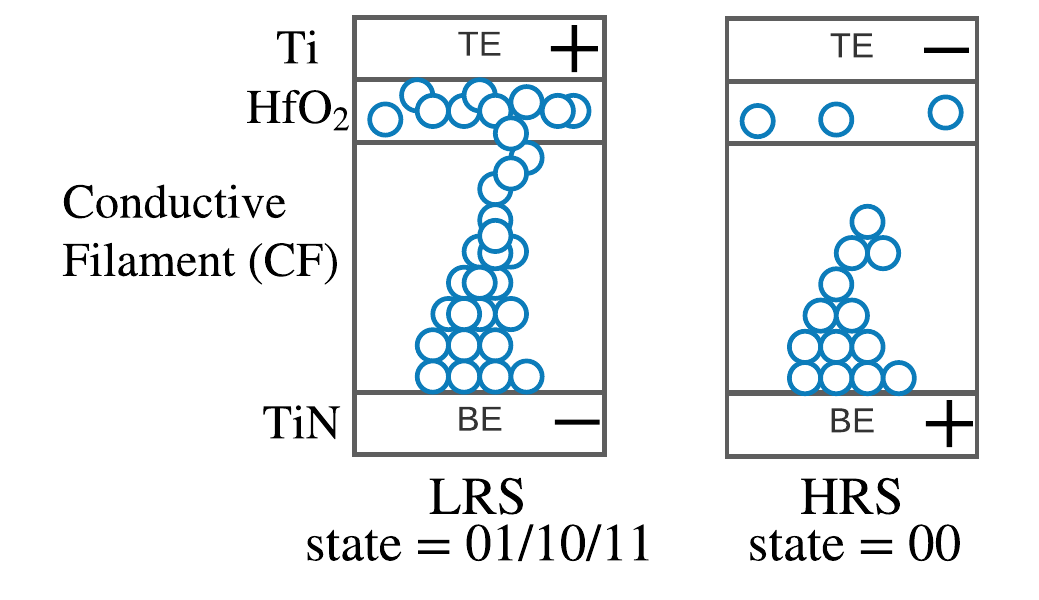}
		\vspace{-10pt}
		\caption{Operation of an RRAM cell with the $\text{HfO}_2$ layer sandwiched between the metals Ti (top electrode) and TiN (bottom electrode). The left subfigure shows the formation of LRS states with the formation of conducting filament (CF). This represents logic states 01, 10, and 11. The right subfigure shows the depletion of CF on application of a negative voltage on the TE. This represents the HRS state or logic 00.}
		\label{fig:RRAM}
		\vspace{-10pt}
	\end{center}
\end{figure}

The resistance switching random access memory (RRAM) technology presents an attractive option for implementing the synaptic cells of a crossbar due to its demonstrated potential for low-power multilevel operation and high integration density~\cite{mallik2017design}. An RRAM cell is composed of an insulating film sandwiched between conducting electrodes forming a metal-insulator-metal (MIM) structure (see Figure~\ref{fig:RRAM}). Recently, filament-based metal-oxide RRAM implemented with transition-metal-oxides such as HfO${}_2$, ZrO${}_2$, and TiO${}_2$ has received considerable attention due to their low-power and CMOS-compatible scaling.

Synaptic weights are represented as conductance of the insulating layer within each RRAM cell. To program an RRAM cell, elevated voltages are applied at the top and bottom electrodes, which re-arranges the atomic structure of the insulating layer. Figure~\ref{fig:RRAM} shows the High-Resistance State (HRS) and the Low-Resistance State (LRS) of an RRAM cell. An RRAM cell can also be programmed into intermediate low-resistance states, allowing its multilevel operations. In this work, we consider each RRAM cell to be programmed to one HRS and three LRS states, implementing two bits per synapse.

\begin{figure*}[h!]%
    \centering
    \subfloat[65nm, 25$^\circ$C.\label{fig:65_25}]{{\includegraphics[width=4.7cm]{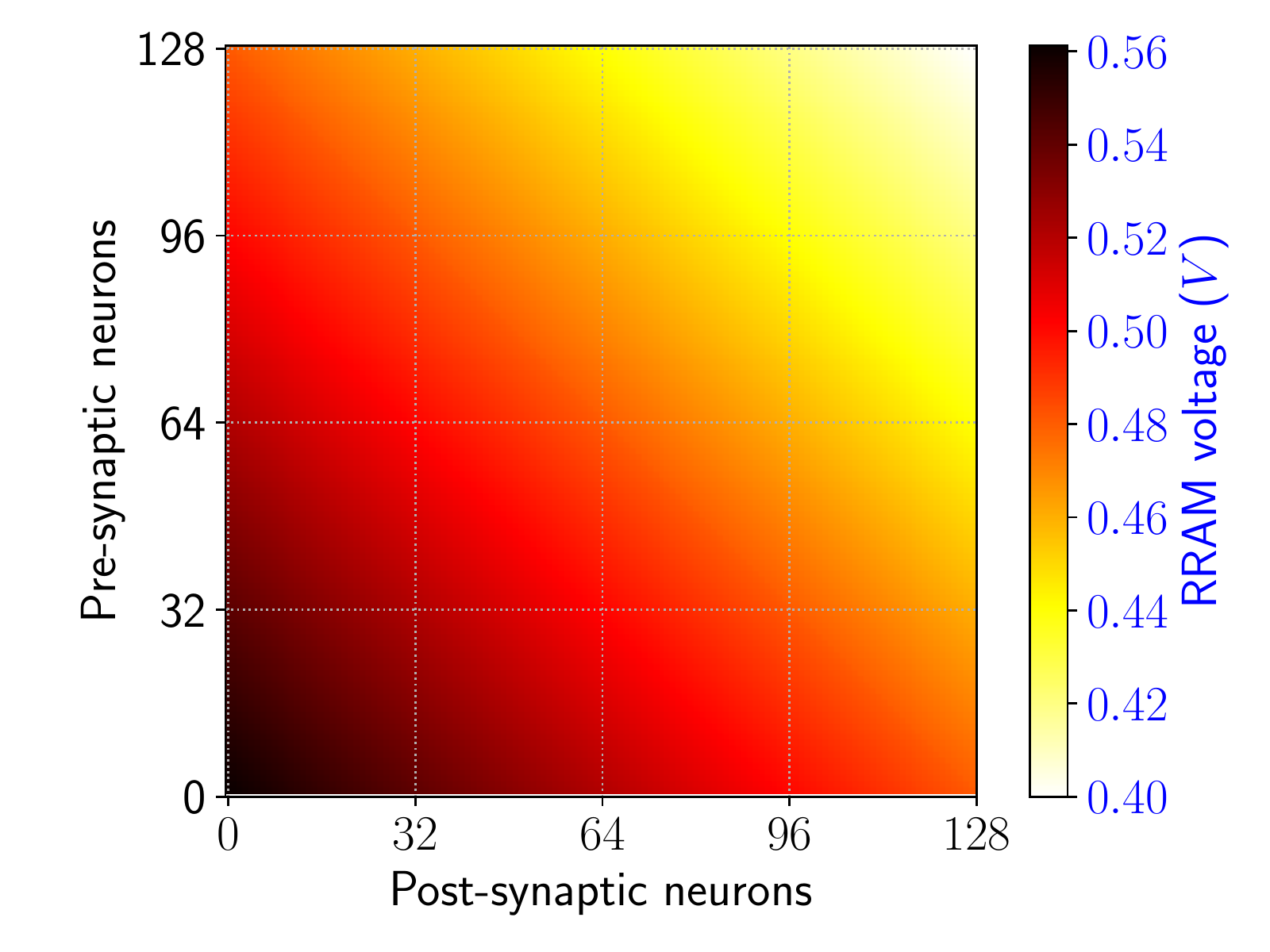} }}%
    \subfloat[45nm, 25$^\circ$C.]{{\includegraphics[width=4.7cm]{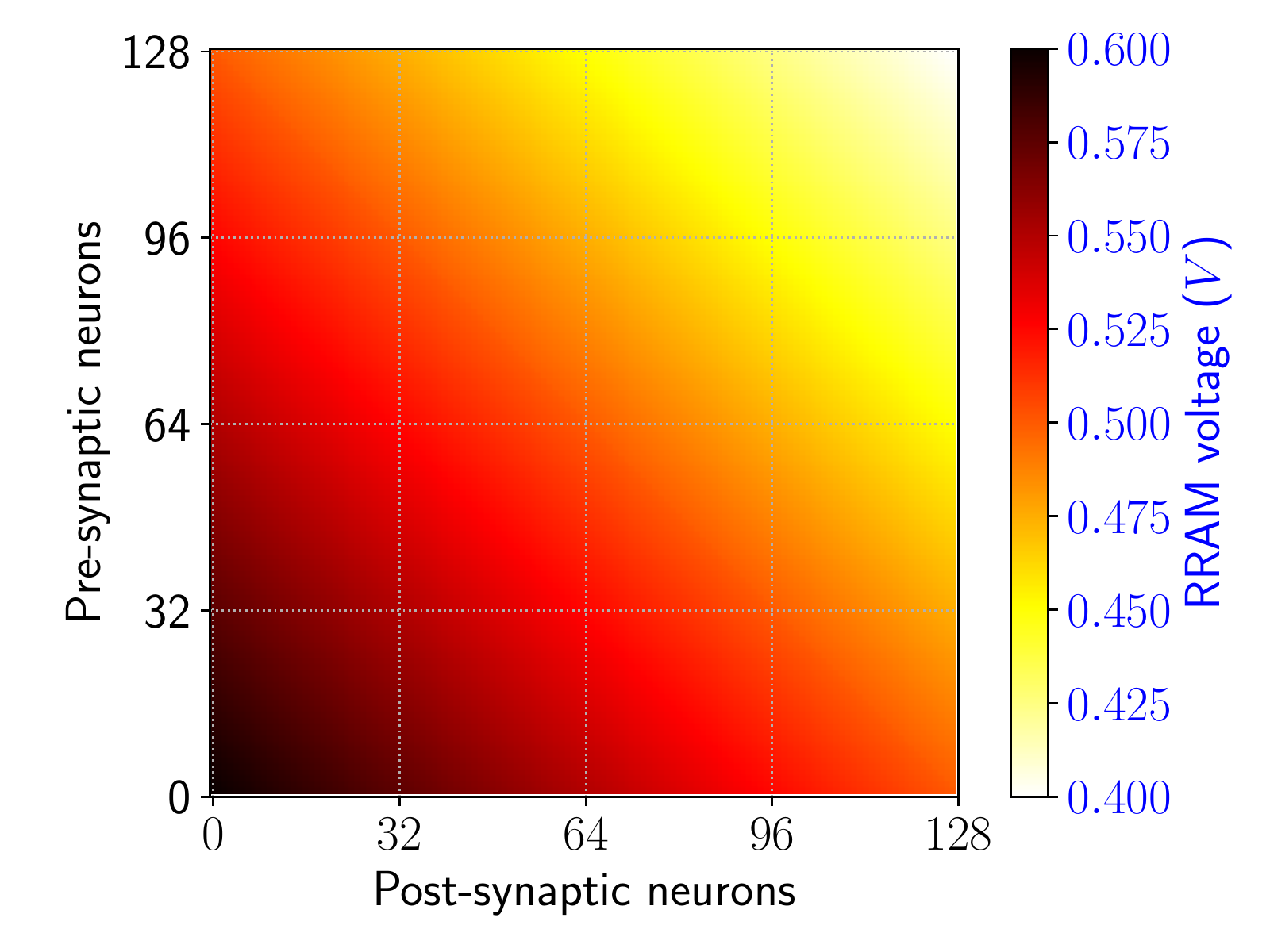} }}%
    \subfloat[32nm, 25$^\circ$C.]{{\includegraphics[width=4.7cm]{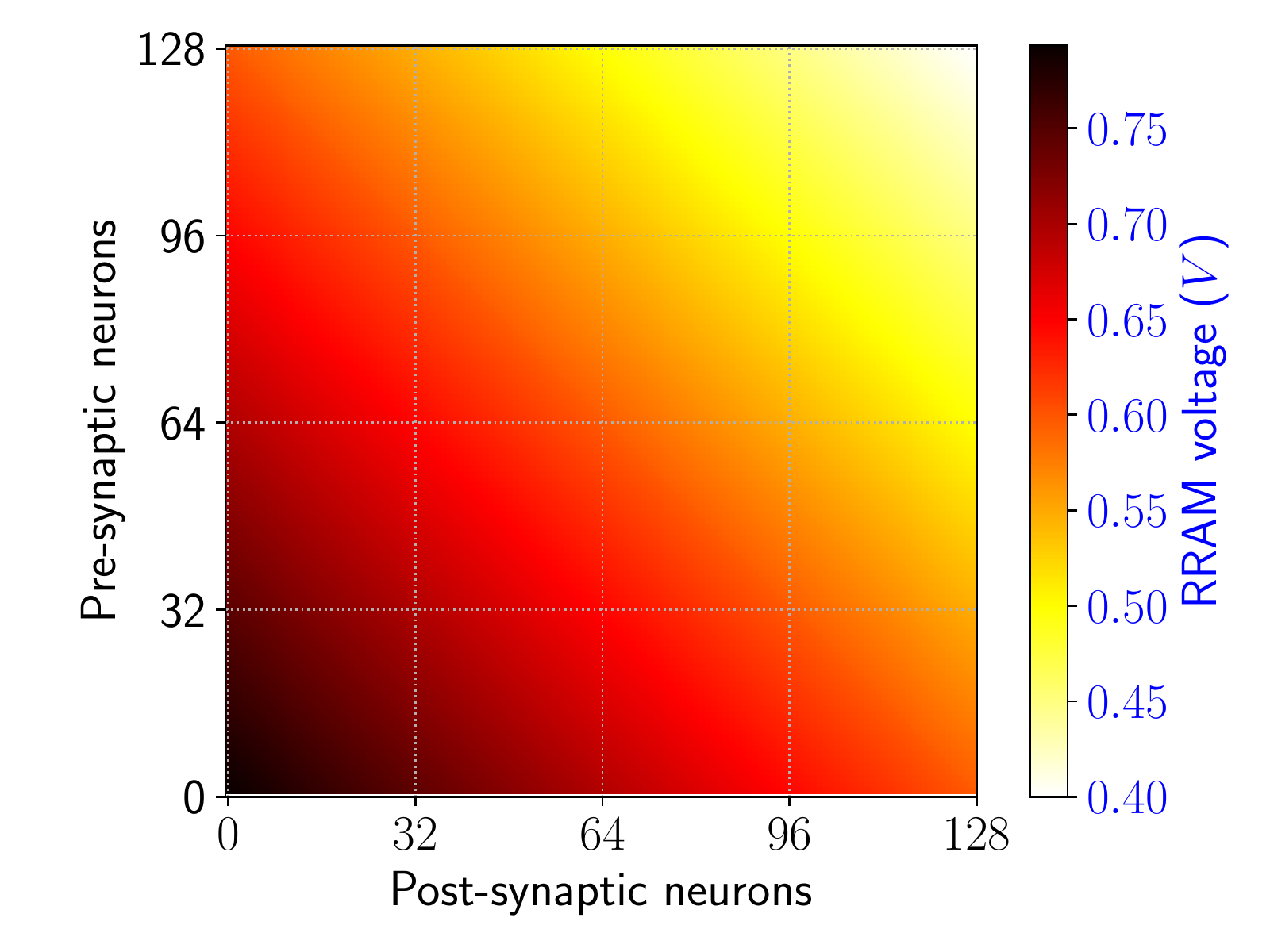} }}%
    \subfloat[16nm, 25$^\circ$C.\label{fig:16_25}]{{\includegraphics[width=4.7cm]{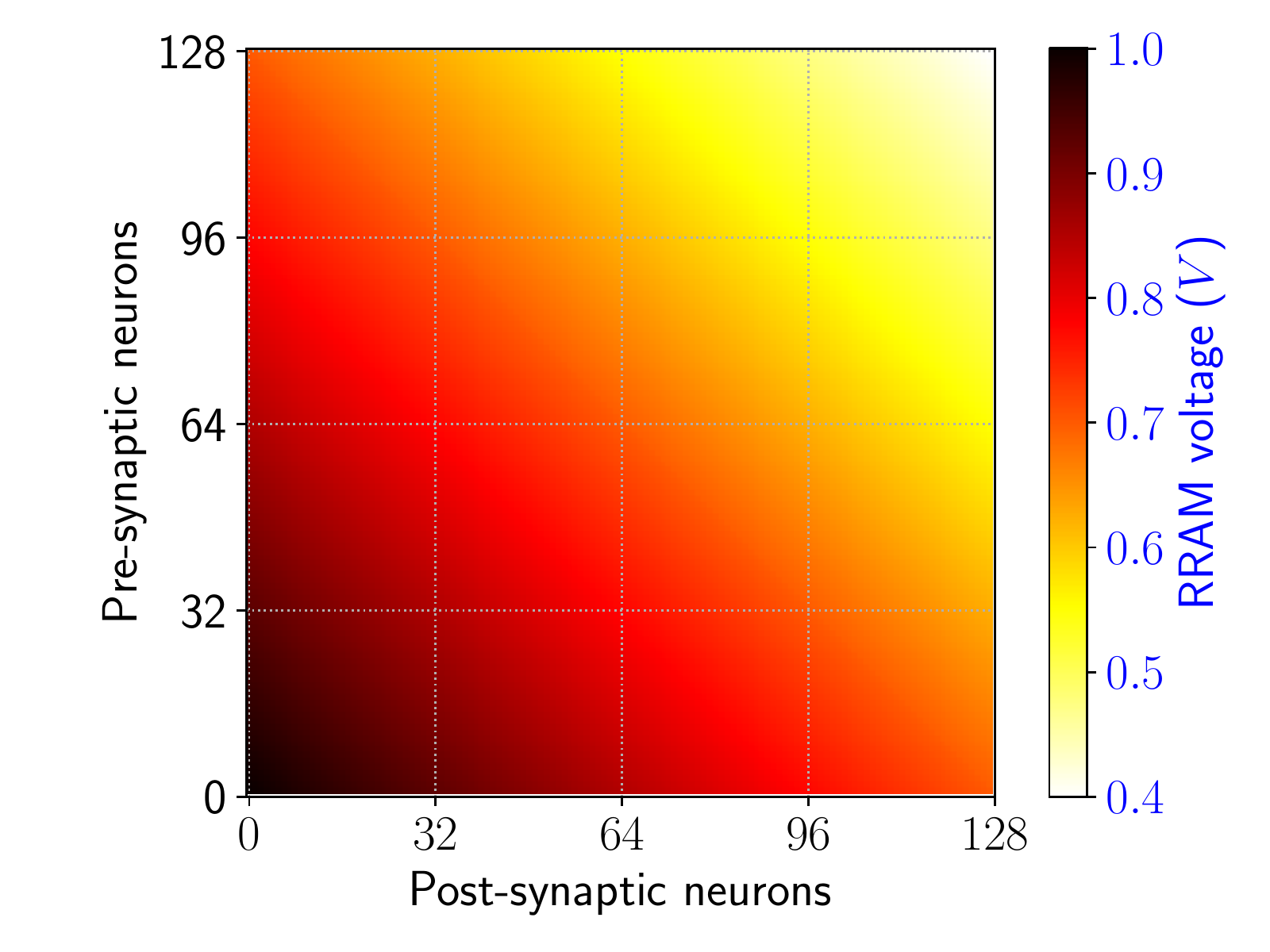} }}%
    \quad
    \subfloat[65nm, 50$^\circ$C.\label{fig:65_50}]{{\includegraphics[width=4.7cm]{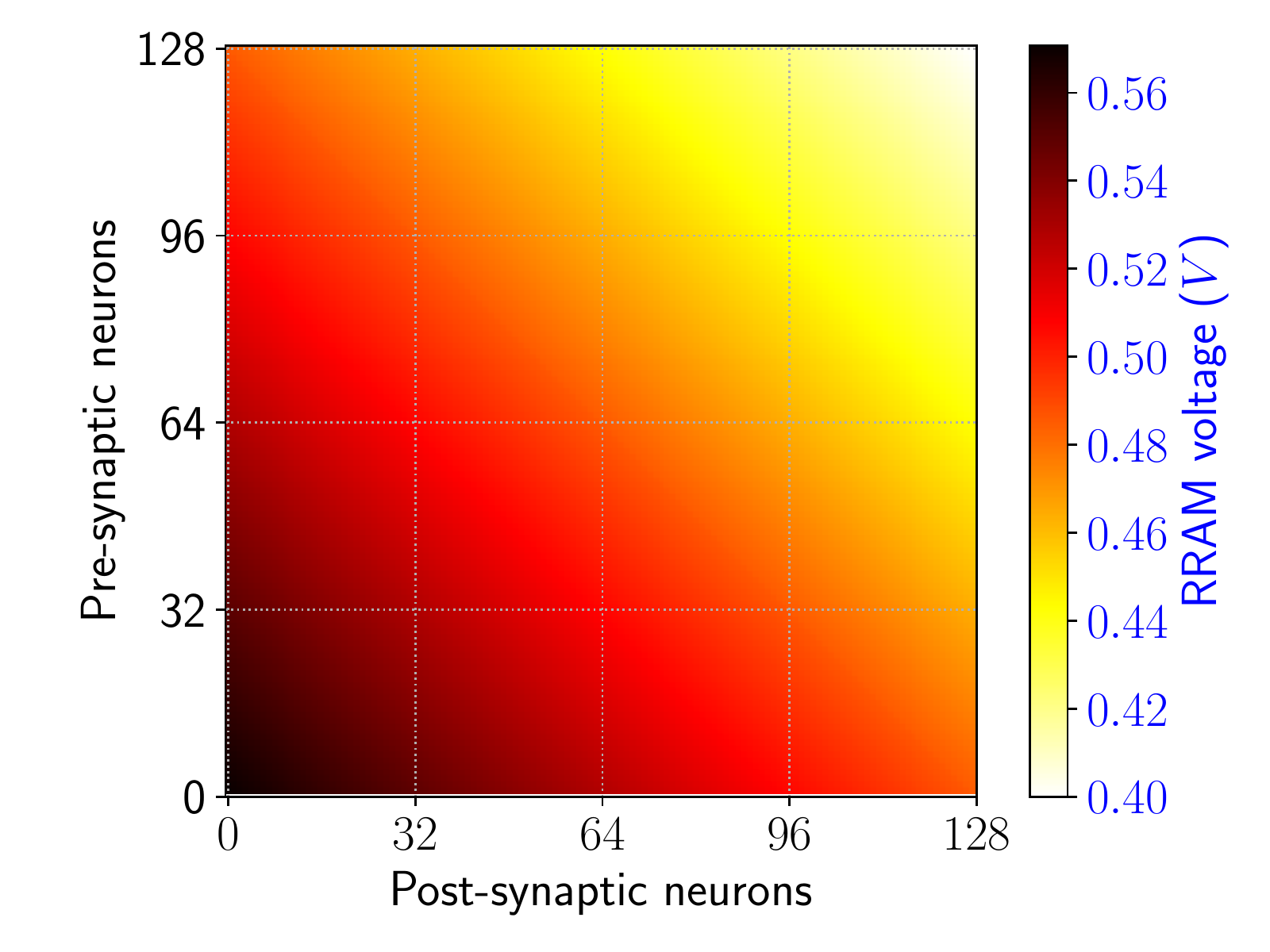} }}%
    \subfloat[45nm, 50$^\circ$C.]{{\includegraphics[width=4.7cm]{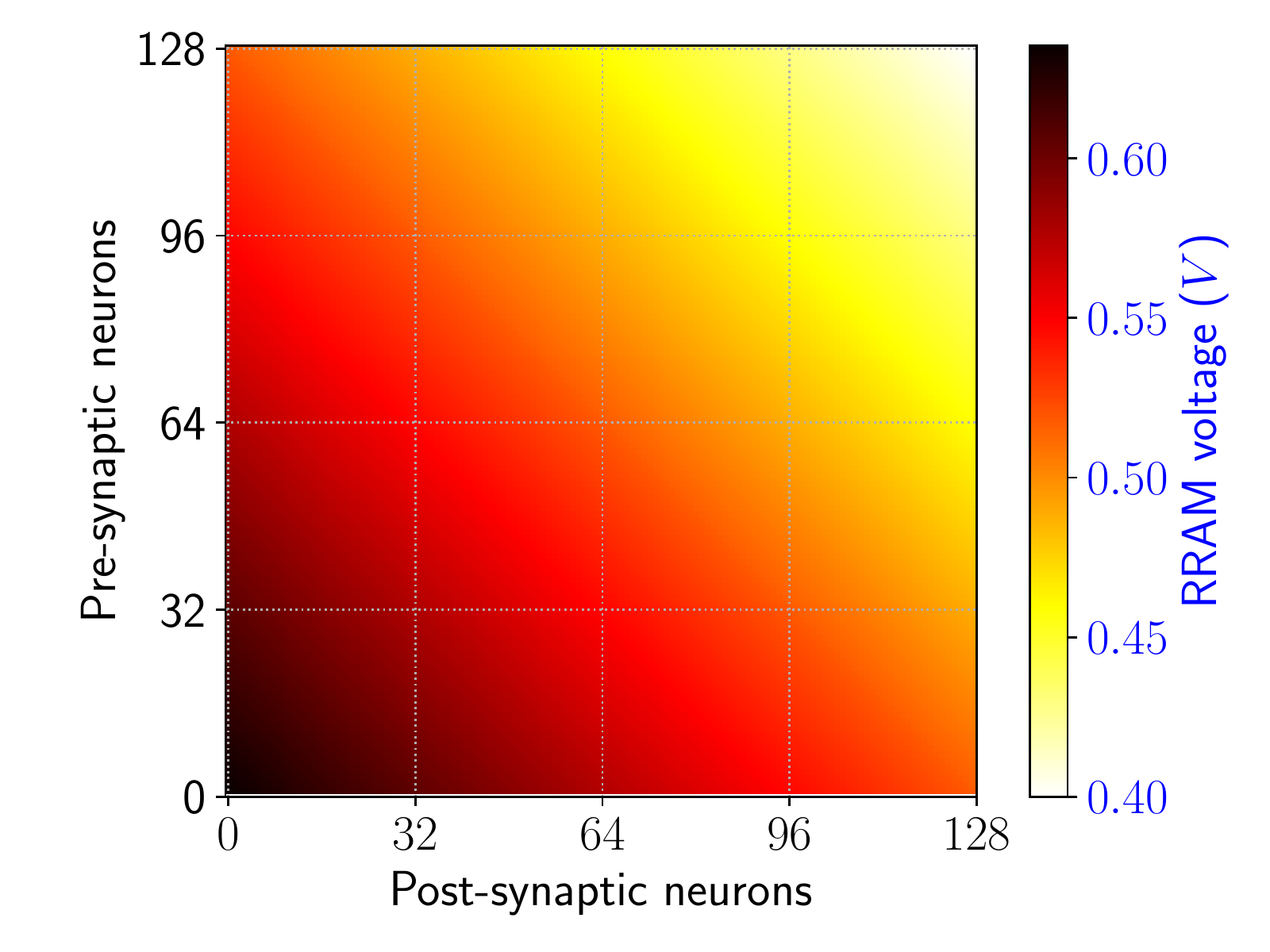} }}%
    \subfloat[32nm, 50$^\circ$C.]{{\includegraphics[width=4.7cm]{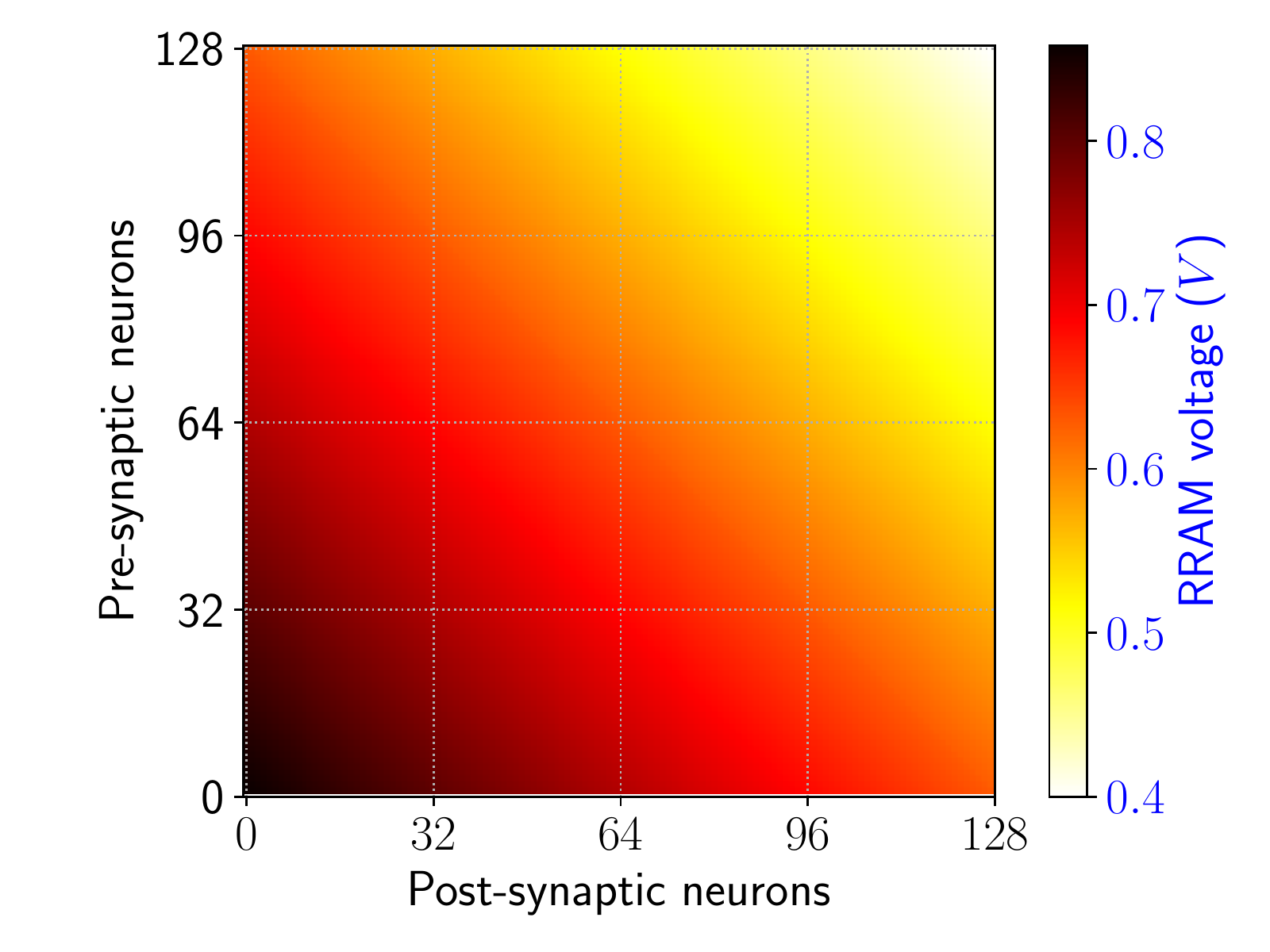} }}%
    \subfloat[16nm, 50$^\circ$C.\label{fig:16_50}]{{\includegraphics[width=4.7cm]{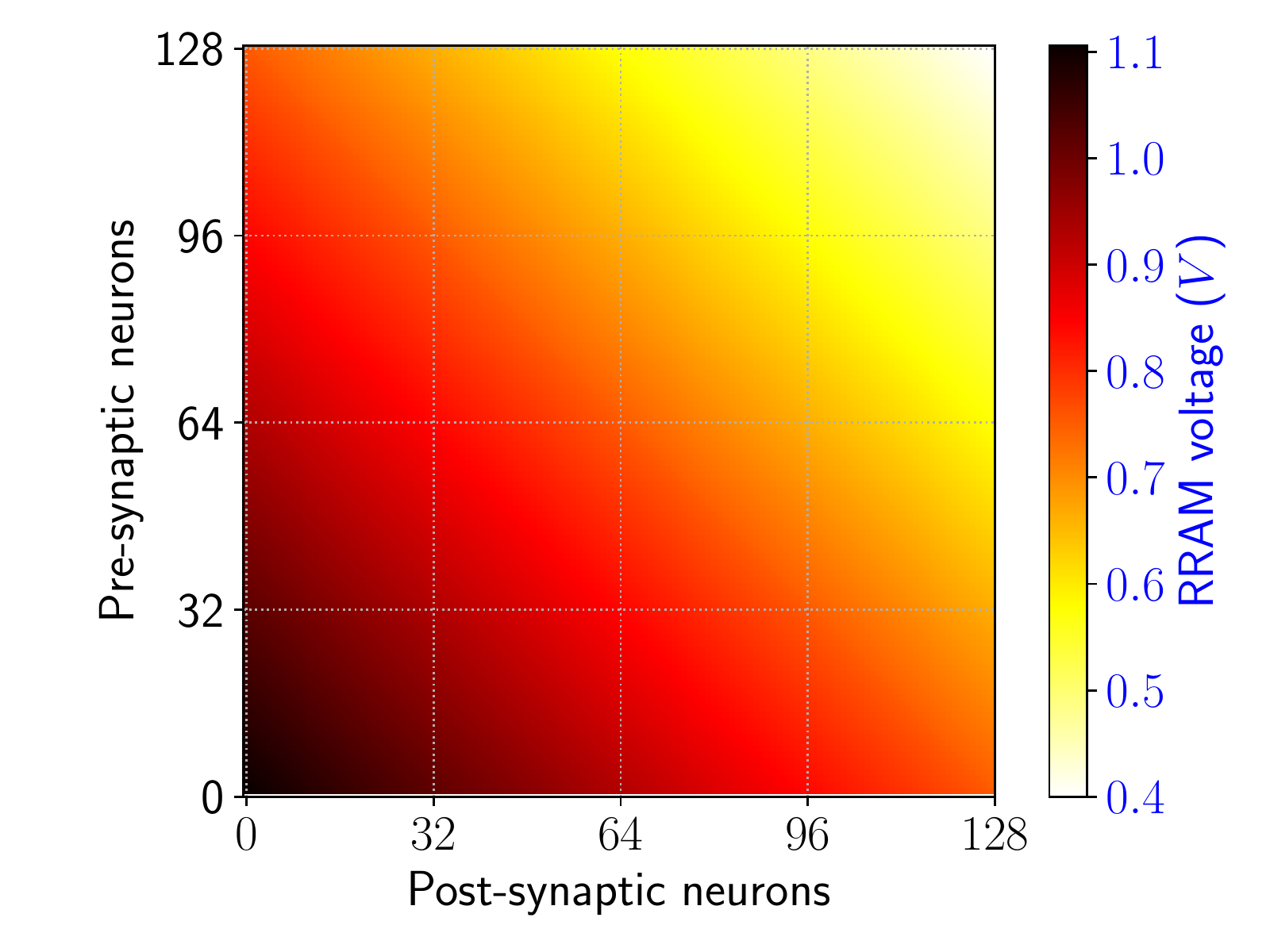} }}%
    \caption{Variation in RRAM voltages within an 128x128 crossbar for inference. Such variations are reported for four technology nodes (65nm, 45nm, 32nm, and 16nm) and two temperature settings (25$^\circ$C and 50$^\circ$C).}%
    \label{fig:pvt_sim}%
\end{figure*}

%% file: sections/endurance_variation.tex
Inside a neuromorphic system, the long bitlines and wordlines of a crossbar are the major sources of parasitic (IR) voltage drops, introducing asymmetry in the voltage applied across the different RRAM cells in the hardware~\cite{hebe,shihao_igsc,mneme}.
To study this behavior, we simulate a 128x128 RRAM-based crossbar circuit using the predictive technology model (PTM)~\cite{zhao2007predictive} and RRAM-specific parameters~\cite{chen2015compact}.

Figure~\ref{fig:pvt_sim} shows the variation of RRAM voltages in a 128x128 crossbar during inference for four technology nodes (65 nm, 45 nm, 32 nm, and 16 nm) and two temperature settings (25$^\circ$C and 50$^\circ$C). We make the following three key observations. 
First, RRAM voltages at the bottom left corner of the crossbar are higher than those at the top right corner. This is because the current paths via the RRAM cells at the bottom left corner are shorter, i.e., they have lower parasitic voltage drops than at the top right corner.
Second, with technology scaling, the voltage variation increases. The highest RRAM voltage in the crossbar is 1.1~V at 16~nm and 25$^\circ$C (Figure~\ref{fig:16_25}) compared to 0.57~V at 65~nm and 25$^\circ$C (Figure~\ref{fig:65_25}). This difference is because the unit parasitic resistance of the electrodes increases from \ineq{1\Omega} at 65~nm to \ineq{3.8\Omega} at 16~nm~\cite{zhao2007predictive}. The value of the parasitic resistance is expected to increase with technology scaling, with a value \ineq{\approx 25\Omega} at {5~nm}~\cite{fouda2017modeling}. 
Third, RRAM voltage increases with temperature (Figures~\ref{fig:65_50}-\ref{fig:16_50} vs. Figures~\ref{fig:65_25}-\ref{fig:16_25}). This is because with increase in temperature, the leakage current through the access transistor of each RRAM cell in the crossbar increases. Therefore, to obtain a certain current margin at the readout unit of the crossbar, the input voltage applied on the top electrodes needs to increase, which increases the RRAM voltages.

\subsection{Voltage-Dependant Read Endurance of RRAM cells}
We now provide a formulation of the read endurance of the RRAM cells in a crossbar as a function of the input voltage.\footnote{Limited write endurance of RRAM cells has been studied before in the context of neuromorphic computing~\cite{espine,twisha_endurance}. This is the first work that studies the read endurance problem and proposes an intelligent solution.} In RRAM technology, the transition from HRS state is governed by a sudden decrease of the vertical filament gap on application of stress voltage during spike propagation~\cite{shim2020impact}. The rate of change of the filament gap of the RRAM cell at the \ineq{(i,j)^\text{th}} location in the crossbar is
\begin{equation}
    \label{eq:hrs}
    \footnotesize \frac{dg_{i,j}}{dt} = -\vartheta_0\cdot e^{-\frac{E_a}{kT}}sinh\left(\frac{\gamma_{i,j}\cdot a_0}{L}\cdot\frac{qV_{i,j}}{kT}\right) \text{, where } \gamma_{i,j} = \gamma_0 - \beta\cdot\frac{g_{i,j}}{g_0}^3
\end{equation}
In the above equation, \ineq{t} defines the state transition time, \ineq{g_0} is the initial filament gap of the RRAM cell, \ineq{V_{i,j}} is the voltage applied to the cell, \ineq{\gamma_{i,j}} is the local field enhancement factor and is related to the gap \ineq{g_{i,j}}, \ineq{a_0} is the atomic hoping distance, and \ineq{\gamma_0} is a fitting constant.

The transition from one of the LRS states is governed by the lateral filament growth~\cite{shim2020impact}. The time for state transition in the \ineq{(i,j)^\text{th}} RRAM cell is given by 
\begin{equation}
    \label{eq:lrs}
    \footnotesize t_{i,j} (LRS) = 10^{-14.7\cdot V_{i,j} + 6.7} \text{sec}
\end{equation}

Using Equations~\ref{eq:hrs} and \ref{eq:lrs}, the read endurance of the \ineq{(i,j)^\text{th}} RRAM cell can be derived as 
\begin{equation}
    \label{eq:endurance}
    \footnotesize E_{i,j} (HRS/LRS) = \frac{t_{i,j} (HRS/LRS)}{1~ms \text{ (spike duration)}}
\end{equation}

\begin{figure}[h!]%
    \centering
    \subfloat[Each RRAM cell in HRS state.\label{fig:hrs_var}]{{\includegraphics[width=4.7cm]{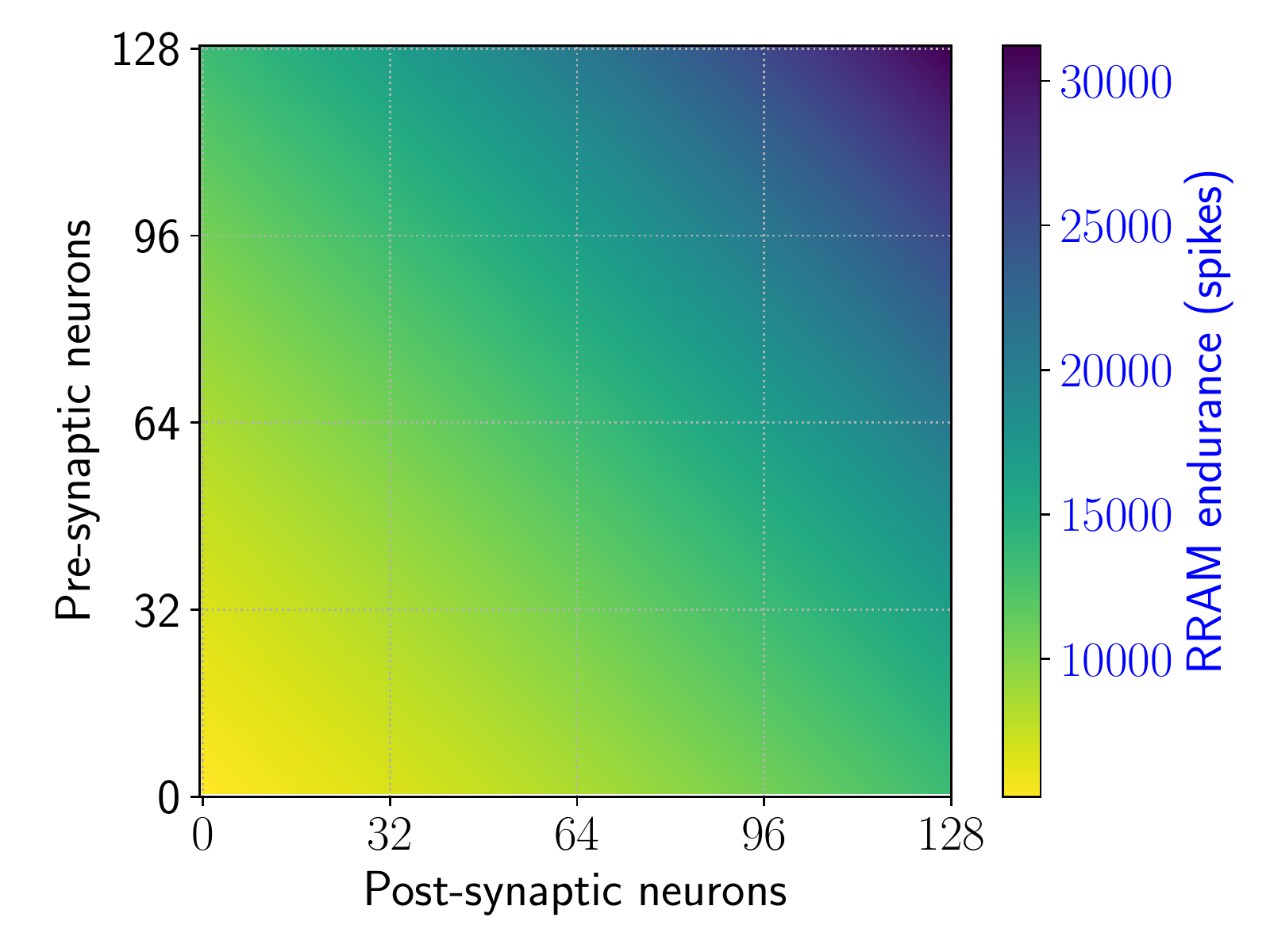} }}%
    \subfloat[Each RRAM cell in LRS state.\label{fig:lrs_var}]{{\includegraphics[width=4.7cm]{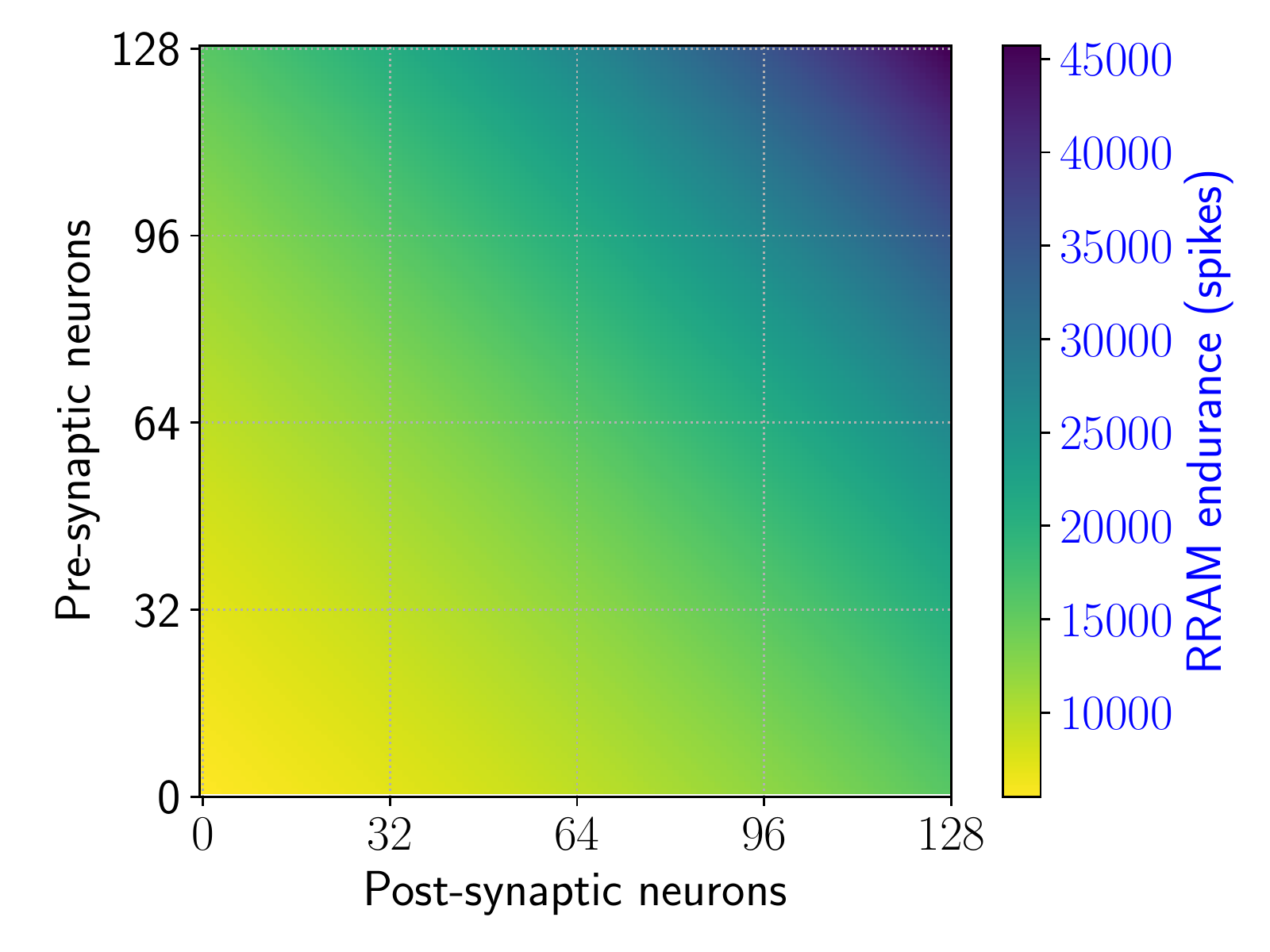} }}%
    \vspace{-5pt}
    \caption{Variation in RRAM endurance within an 128x128 crossbar at 45~nm node and at 25$^\circ$C. Such variations are reported for the RRAM cells programmed in (a) HRS and (b) one of the LRS states.}%
    \label{fig:endurance_map}%
\end{figure}

Figure~\ref{fig:endurance_map} shows the endurance variation of a 128x128 crossbar at 45~nm node and at 25$^\circ$C with each RRAM cell programmed to 1) HRS state (Figure~\ref{fig:hrs_var}) and 2) one of the LRS states (Figure~\ref{fig:lrs_var}). We observe that endurance of an RRAM cell is higher if the cell is programmed to an LRS state compared to  when it is programmed to the HRS state.

From the machine learning workload perspective, synapses can either be in the HRS state or in one of the LRS states. Therefore, based on how the synapses of a model are mapped inside a crossbar, the endurance map will assume intermediate forms between Figures~\ref{fig:hrs_var} and \ref{fig:lrs_var}.
To simplify the problem formulation, we define \textit{equivalence} in terms of inference lifetime as follows. Consider \ineq{A} and \ineq{B} to be two synaptic weights in a machine learning model with spikes \ineq{S_A} and \ineq{S_B}, respectively. Without loss of generality, consider \ineq{A} to be in HRS state and \ineq{B} in LRS state. Let these weights are programmed on the RRAM cells at the same \ineq{(i,j)^\text{th}} location in two different crossbars. Then, the inference lifetime due to \ineq{A} is \ineq{\frac{E_{i,j}(HRS)}{S_A}} and that due to \ineq{B} is \ineq{\frac{E_{i,j}(LRS)}{S_B}}. Synaptic weights \ineq{A} and \ineq{B} are considered to be equivalent in terms of the inference lifetime at the \ineq{(i,j)^\text{th}} position in a crossbar if 
\begin{equation}
    \label{eq:lifetime_equivalence}
    \footnotesize \frac{E_{i,j}(HRS)}{S_A} = \frac{E_{i,j}(LRS)}{S_B}
\end{equation}

We use Equation~\ref{eq:lifetime_equivalence} 
for mapping and inference lifetime computation purposes.
Once the mapping is decided, RRAM cells are programmed to their actual state.

%% file: sections/problem_formulation.tex
The mapping of a machine learning model to hardware is formulated in the following three steps.
\begin{enumerate}
    \item Formulating inference lifetime of model clusters.
    \item Cluster mapping with unlimited hardware resources.
    \item Cluster mapping with limited hardware resources.
\end{enumerate}
We now elaborate on these mapping steps.
\subsection{Formulating Inference Lifetime of Model Clusters}\label{sec:endurance_formulation}
We consider the mapping of a cluster \ineq{\mathbb{C} = (Pre, Post, Syn)} of a machine learning model onto a crossbar of the hardware. Here \ineq{Pre} is the set of pre-synaptic neuron, \ineq{Post} is the set of post-synaptic neuron, and \ineq{Syn} is the set of synapses between the pre- and post-synaptic neurons of the cluster. Each neuron \ineq{n_i\in Pre} is characterized by a number \ineq{spk(n_i)}, indicating the average number of spikes generated by this neuron per image during inference. Each synapse \ineq{s_{i,j}\in Syn} connecting the pre-synaptic neuron \ineq{n_i\in Pre} and post-synaptic neuron \ineq{n_j\in Post} is associated with a number \ineq{wt(s_{i,j})} representing its synaptic weight. The number of spikes on the synapse \ineq{s_{i,j}} is the same as the number of spikes generated by its pre-synaptic neuron \ineq{n_i}, i.e., \ineq{spk(s_{i,j}) = spk(n_i)}.

Consider the mapping of this cluster \ineq{\mathbb{C}} to a MxM crossbar \ineq{\mathbb{H} = (In,Out)} with a set \ineq{In} of input ports to map pre-synaptic neurons and a set \ineq{Out} of output ports to map post-synaptic neurons. Here \ineq{|In| = |Out| = M}. 

Let \ineq{X_{i,k}} be a binary variable representing the mapping of pre-synaptic neuron \ineq{n_i\in Pre} to the input port \ineq{i_k\in In} and \ineq{Y_{j,l}} be a binary variable representing the mapping of post synaptic neuron \ineq{n_j\in Post} to the output port \ineq{o_l\in Out}.

The problem we are aiming to solve is this: find the binary variables \ineq{X_{i,j}} and \ineq{Y_{j,l}} such that the inference lifetime is maximized when mapping the cluster to a crossbar. Therefore, the \textit{objective function} is the inference lifetime. To maximize the objective function, we define the following constraints.
\begin{itemize}
    \item Each pre-synaptic neuron can be mapped to exactly one input port of the hardware, i.e., \ineq{\sum_{k=1}^M X_{i,k} = 1~\forall i}.
    \item Each post-synaptic neuron can be mapped to exactly one output port of the hardware, i.e., \ineq{\sum_{l=1}^M Y_{j,l} = 1~\forall j}
\end{itemize}

To formulate the objective function itself, we consider the synapse \ineq{s_{i,j} \in Syn}, which connects the pre-synaptic neuron \ineq{n_i} with the post-synaptic neuron \ineq{n_j}. In terms of the variables \ineq{X_{i,k}} and \ineq{Y_{j,l}}, the synapse \ineq{s_{i,j}} is mapped to the RRAM cell at \ineq{(k,l)^\text{th}} position in the crossbar. The inference lifetime of the synapse can be computed by first considering the equivalence to HRS state using Equation~\ref{eq:lifetime_equivalence}, and then dividing the HRS endurance of the RRAM cell with the number of spikes on the synapse using Equation~\ref{eq:lifetime_computation}. This is given by
\begin{equation}
    \label{eq:lifetime_formulation}
    \footnotesize f(i,j) = \text{Inference Lifetime}(i,j) =  \sum_{k=1}^M\sum_{l=1}^M X_{i,k}\cdot Y_{j,l}\cdot\frac{E_{k,l}(HRS)}{spk_{eq}(s_{i,j})}
\end{equation}

The maximization problem is
\begin{equation}
    \label{eq:maximize_objective_fn}
    \footnotesize \max_{\substack{1 \leq i \leq |Pre| \\ 1 \leq j \leq |Post|}}  f(i,j)
\end{equation}

The use of binary (discrete) variables makes the optimization problem of Equation~\ref{eq:maximize_objective_fn} non-convex, while the product term in Equation~\ref{eq:lifetime_formulation} makes this non-linear (NL). Therefore, the optimization problem we are aiming to solve is a Non-Convex Binary Non-Linear Programming (BNLP) problem and there is no guarantee of optimality~\cite{boyd2004convex}. We use the smoothing method proposed in the Ph.D. dissertation~\cite{ng2002continuation} to solve this BNLP problem. The optimized inference lifetime obtained from mapping the cluster \ineq{\mathbb{C}} to the crossbar \ineq{\mathbb{H}} is 
\begin{equation}
    \label{eq:optimized_lifetime}
    \footnotesize \text{Inference Lifetime} (\mathbb{C},\mathbb{H}) =  f_\text{opt}
\end{equation}

In this study, we have ignored process variations across the crossbars of a hardware. So, the inference lifetime of a cluster is the same, irrespective of which crossbar this cluster is mapped to in the hardware. This allows us to decouple the crossbar term from Equation~\ref{eq:optimized_lifetime}.
We use the variable \ineq{\mathcal{L}_i} to represent the inference lifetime of the cluster \ineq{\mathbb{C}_i}.


\subsection{Cluster Mapping with Unlimited Hardware Resources}\label{sec:mapping_unlimited}
If the hardware contains unlimited number of crossbars, then 
each crossbar can map at most one cluster of a machine learning model.
Therefore, the inference lifetime for the model when it is mapped to the hardware is the minimum inference lifetime of all the clusters of the hardware, i.e,
\begin{equation}
    \label{eq:inference_lifetime_unlimited}
    \footnotesize \text{Inference Lifetime} = \text{minimum}\{\mathcal{L}_i,~\forall i \in 1, 2, \cdots, N_C\},
\end{equation}
where \ineq{N_C} is the number of clusters of the model.

\subsection{Cluster Mapping with Limited Hardware Resources}\label{sec:mapping_limited}
If the number of crossbars in the hardware is limited, then each crossbar may need to be shared across multiple clusters of a machine learning model. We now formulate the cluster mapping problem as follows. Let the binary variable \ineq{Z_{i,j}} indicate the mapping of cluster \ineq{\mathbb{C}_i} to crossbar \ineq{\mathbb{H}_j}, i.e.,
\begin{equation}
    \footnotesize Z_{i,j} = \begin{cases}
    1 & \text{ if cluster } \mathbb{C}_i \text{ is mapped to crossbar } \mathbb{H}_j\\
    0 & \text{otherwise}
    \end{cases}
\end{equation}

The problem we are aiming to solve is this: find the binary variables \ineq{Z_{i,j}} such the inference lifetime of the machine learning model on the hardware is improved. We define the following constraint: each cluster must be mapped to only one crossbar, i.e., \ineq{\sum_{j=1}^{N_H} Z_{i,j} = 1~\forall i},
where \ineq{N_H} is the number of crossbars of the hardware with \ineq{N_H \leq N_C}. 

\begin{figure}[h!]
	\begin{center}
		\vspace{-10pt}
		\includegraphics[width=0.99\columnwidth]{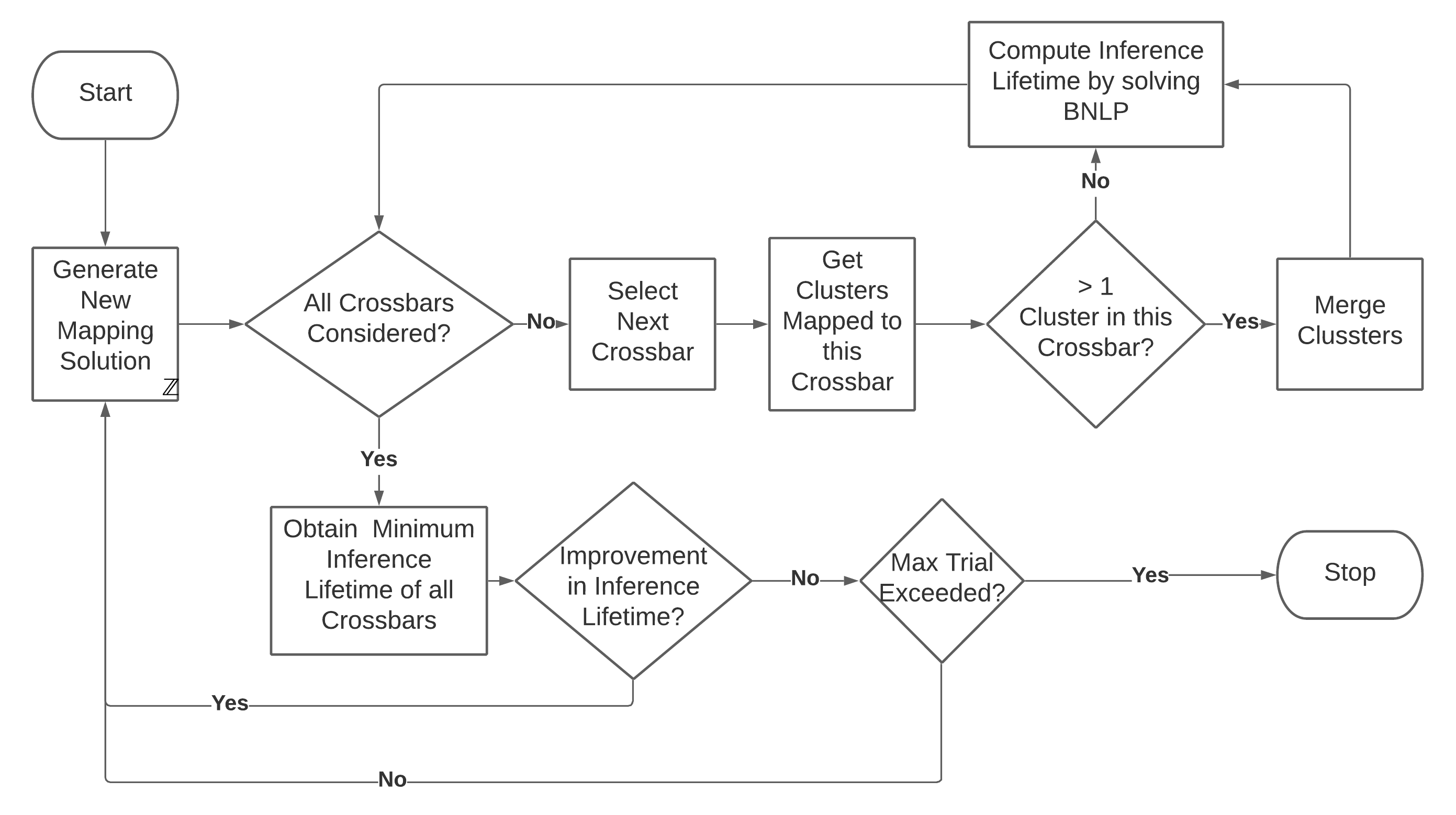}
		\vspace{-15pt}
		\caption{Flowchart for Hill-Climbing-based search.}
		\label{fig:hill_climbing_search}
		\vspace{-15pt}
	\end{center}
\end{figure}

To explore the cluster-to-crossbar mapping search space for the maximum inference lifetime, we use a \textit{Hill-Climbing}-based local search~\cite{selman2006hill}. Each mapping solution is represented by \ineq{\mathbb{Z}\in \mathbb{R}^{N_C\times N_H}}. Figure~\ref{fig:hill_climbing_search} shows the working of the search algorithm.

For each solution \ineq{\mathbb{Z}} generated by the algorithm, it computes the inference lifetime of each crossbar as follows. If a crossbar has only one cluster, the inference lifetime is computed by solving the proposed BNLP problem formulation (Equation~\ref{eq:maximize_objective_fn}). If more than one cluster is mapped to the crossbar, the clusters on this crossbar are merged. Merging a cluster involves combining the clusters to form a larger cluster and is described mathematically as follows. Merging of clusters \ineq{\mathbb{C}_a =  (Pre_a,Post_a,Syn_a)} and \ineq{\mathbb{C}_b = (Pre_b,Post_b,Syn_b)} is
\begin{equation}
    \footnotesize \texttt{merge}(\mathbb{C}_a,\mathbb{C}_b) = \mathbb{C}_{a+b} = (Pre_{a+b},Post_{a+b},Syn_{a+b}),
\end{equation}
where \ineq{Pre_{a+b} = Pre_a\cup Pre_b}, \ineq{Post_{a+b} = Post_a\cup Post_b}, and \ineq{Syn_{a+b} = Syn_a\cup Syn_b}.
Once the clusters mapped to a crossbar are merged, the inference lifetime of the merged cluster is computed using the proposed BNLP formulation. 
Then the algorithm computes the overall inference lifetime as the minimum of the inference lifetime of all crossbars of the hardware. If this value is higher than the best solution obtained thus far, the mapping is retained and the algorithm proceeds to find a better mapping solution. Otherwise, the algorithm continues to explore for a few more iterations to see if a better mapping solution can be generated. 
This general formulation can be applied to the case where \ineq{N_H > N_C}, i.e., the number of hardware crossbars is greater than model clusters.

\subsection{Performance Impact}
We now formally quantify the performance degradation due to our mapping exploration (Sections~\ref{sec:endurance_formulation}, \ref{sec:mapping_unlimited}, and \ref{sec:mapping_unlimited}) using the previously-introduced notations. Let \ineq{\mathbb{Z}_\text{opt}\in \{0,1\}^{N_C\times N_H}} be the optimum mapping (one with the highest inference lifetime obtained using the Hill-Climbing approach of Figure~\ref{fig:hill_climbing_search}) of the clusters of a machine learning model to the crossbars of a neuromorphic hardware. 

Within the optimum mapping, let \ineq{\mathbb{M}_{\text{opt}_p} = [\mathbb{X}_{\text{opt}_p}~|~\mathbb{Y}_{\text{opt}_p}]~\forall p\in 1,2,\cdots,N_C} be the optimum mapping (one with the highest inference lifetime obtained by solving the BNLP of Equation~\ref{eq:maximize_objective_fn}) of pre- and post-synaptic neurons of the \ineq{p^\text{th}} cluster to a crossbar. Here \ineq{\mathbb{X}_{\text{opt}_p} \in \{0,1\}^{|Pre_p|\times |In|}} and \ineq{\mathbb{Y}_{\text{opt}_p} \in \{0,1\}^{|Post_p|\times |Out|}}. 

Let \ineq{d_{k,l}} represents the delay in spike propagation through the RRAM cell at the \ineq{(k,l)^\text{th}} location in a crossbar. Therefore, the spike propagation delay through the synapse \ineq{s_{i,j}} is
\begin{equation}
    \label{eq:spike_delay_synapse}
    \footnotesize \text{synapse\_delay}_{i,j} = \sum_{k=1}^M\sum_{l=1}^M X_{i,k}\cdot Y_{j,l}\cdot d_{k,l}
\end{equation}
Therefore, the average spike propagation delay of the cluster when mapped to a crossbar is
\begin{equation}
    \label{eq:spike_delay_crossbar}
    \footnotesize \text{cluster\_delay} = \frac{\sum_{i=1}^{Pre}\sum_{j=1}^{Post}spk(s_{i,j})\cdot \text{synapse\_delay}_{i,j}}{\sum_{i=1}^{Pre}\sum_{j=1}^{Post} spk(s_{i,j})} 
\end{equation}
Equation~\ref{eq:spike_delay_crossbar} can be extrapolated to compute the spike propagation delay of the entire machine learning model when mapped to the neuromorphic hardware as
\begin{equation}
    \label{eq:spike_delay_hardware}
    \footnotesize \text{hardware\_delay} = \frac{\sum_{p=1}^{N_C}\sum_{i=1}^{Pre_p}\sum_{j=1}^{Post_p} spk(s_{i,j}^p)\cdot \text{cluster\_delay}_p}{\sum_{p=1}^{N_C}\sum_{i=1}^{Pre_p}\sum_{j=1}^{Post_p} spk(s_{i,j}^p)}
\end{equation}

%% file: sections/results.tex
\begin{figure}[h!]
	\begin{center}
		\vspace{-10pt}
		\includegraphics[width=0.99\columnwidth]{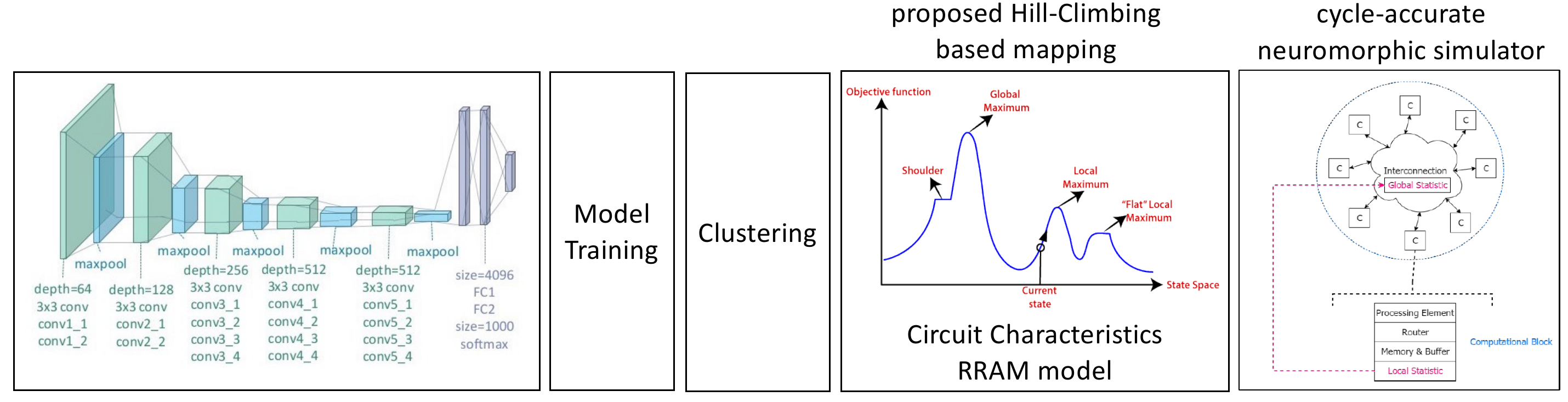}
		\vspace{-15pt}
		\caption{Evaluation framework based on NeuroXplorer~\cite{neuroxplorer}.}
		\label{fig:framework}
		\vspace{-10pt}
	\end{center}
\end{figure}

We evaluate the proposed approach using NeuroXplorer~\cite{neuroxplorer}, a cycle-accurate neuromorphic system simulator that uses tile-based architecture (see Fig.~\ref{fig:framework}). We model the DYNAPs neuromorphic hardware~\cite{dynapse} with hierarchical NoC-based interconnect. Each tile has one $128 \times 128$ crossbar.
Table~\ref{tab:hw_parameters} reports the relevant hardware parameters.

\begin{table}[h!]
    \caption{Major simulation parameters extracted from \cite{dynapse}.}
	\label{tab:hw_parameters}
	\vspace{-5pt}
	\centering
	{\fontsize{6}{10}\selectfont
		\begin{tabular}{lp{5cm}}
			\hline
			Neuron technology & 45nm\\
			\hline
			Synapse technology & {HfO${}_2$-based RRAM}\\
			\hline
			Supply voltage & 1.0V\\
			\hline
	\end{tabular}}
\end{table}

We evaluate 10 machine learning programs which are representative of three most commonly-used neural network classes: convolutional neural network (CNN), multi-layer perceptron (MLP), and recurrent neural network (RNN).
Table~\ref{tab:apps} summarizes the topology, the number of neurons and synapses of these applications, and their baseline accuracy on the DYNAPs neuromorphic hardware using the \sm{}~\cite{spinemap}.

\vspace{-10pt}
\begin{table}[h!]
	\renewcommand{\arraystretch}{0.8}
	\setlength{\tabcolsep}{2pt}
	\caption{Applications used to evaluate the proposed approach.}
	\label{tab:apps}
	\vspace{-5pt}
	\centering
	\begin{threeparttable}
	{\fontsize{6}{10}\selectfont
		\begin{tabular}{ccc|ccl|c}
			\hline
			\textbf{Class} & \textbf{Applications} &
			\textbf{Dataset} &
			\textbf{Synapses} & \textbf{Neurons} & \textbf{Topology} & \textbf{Accuracy}\\
			\hline
			\multirow{4}{*}{CNN} & LeNet & MNIST & 282,936 & 20,602 & CNN & 85.1\%\\
			& AlexNet & ImageNet & 38,730,222 & 230,443 & CNN & 69.8\%\\
			& VGG & CIFAR-10 & 99,080,704 & 554,059 & CNN & 90.7 \%\\
			& HeartClass~\cite{jolpe18,das2018heartbeat} & Physionet & 1,049,249 & 153,730 & CNN & 63.7\%\\
			\hline
			\multirow{3}{*}{MLP} & MLPDigit & MNIST & 79,400 & 884 & FeedForward & 91.6\%\\
			& EdgeDet & CARLsim & 114,057 &  6,120 & FeedForward & 100\%\\
			& ImgSmooth & CARLsim & 9,025 & 4,096 & FeedForward & 100\%\\
			\hline
 			\multirow{3}{*}{RNN} & HeartEstm \cite{HeartEstmNN} & Physionet & 66,406 & 166 & Recurrent Reservoir & 100\%\\
 			& VisualPursuit \cite{Kashyap2018} & \cite{Kashyap2018} & 163,880 & 205 & Recurrent Reservoir & 47.3\%\\
 			& RNNDigit  & MNIST & 11,442 & 567 & Recurrent Reservoir & 83.6\%\\
			\hline
	\end{tabular}}
	\end{threeparttable}
\end{table}
\vspace{-10pt}

\subsection{Inference Lifetime on Unlimited Hardware Resources}
\begin{figure}[h!]
	\begin{center}
		\vspace{-15pt}
		\includegraphics[width=0.99\columnwidth]{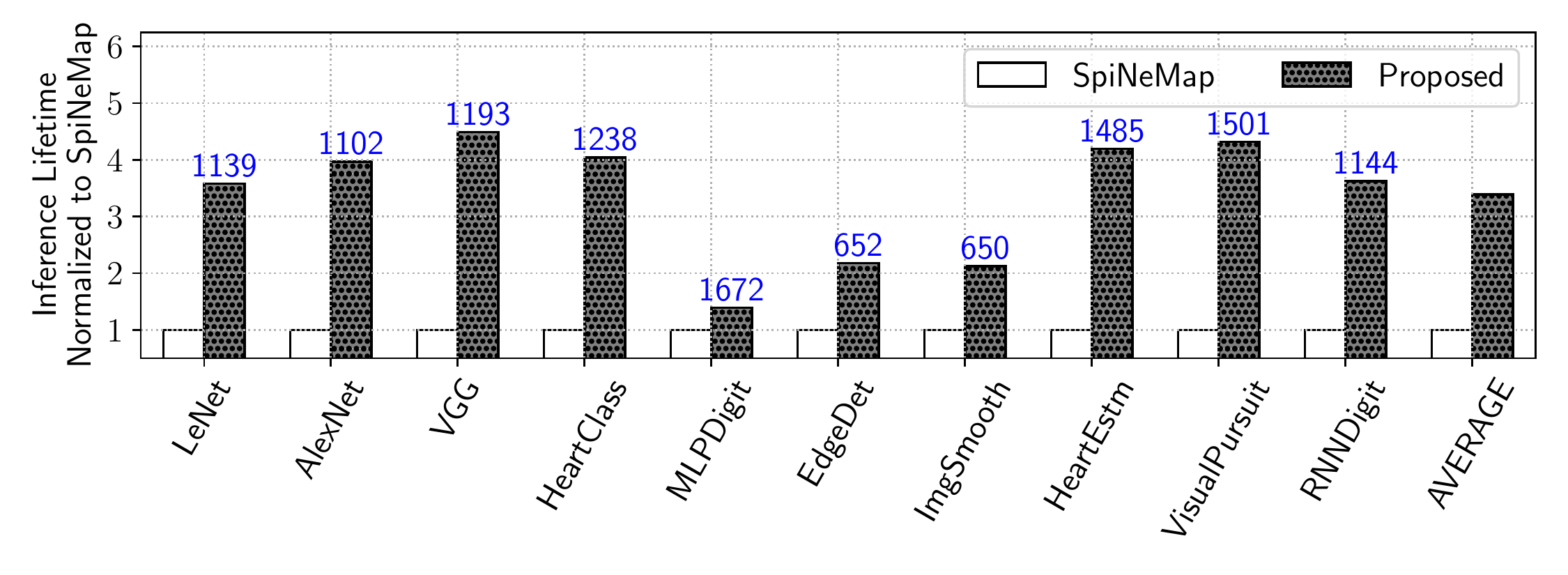}
		\vspace{-25pt}
		\caption{Inference lifetime normalized to SpiNeMap.}
		\vspace{-15pt}
		\label{fig:il_unlimited}
	\end{center}
\end{figure}

Figure~\ref{fig:il_unlimited} reports the inference lifetime for each application for the proposed approach normalized to \sm{}. 
For reference, we have reported the absolute inference lifetime in frames for each application using the proposed approach. For image-based applications (LeNet, AlexNet, VGG, MLPDigit, EdgeDet, ImgSmooth, and RNNDigit), a frame corresponds to an individual image. For other time-series applications (HeartClass, HeartEstm, and VisualPursuit), a frame corresponds to a window of 500ms.
We make the following two observations. 

First, the inference lifetime obtained using the proposed approach is on average 3.4x higher than \sm{}. This improvement is because the proposed approach uses the novel BNLP formulation to decide how the synaptic weights of a cluster need to be mapped to the RRAM cells of a crossbar to maximize the inference lifetime. To do so, the proposed approach incorporates both RRAM device and machine learning workload characteristics. The proposed formulation ensures that critical synapses (those that propagate more spikes) are never mapped to the weaker cells (those that have low read endurance). \sm{} on the other hand, maps the synaptic weight of a cluster arbitrarily to the RRAM cells of a crossbar.

Second, the inference lifetime improvement of the proposed approach is, in general, lower for smaller applications such as MLPDigit, EdgeDet, and ImgSmooth (average 1.9x), compared to larger applications such as LeNet, AlexNet, and VGG (average 4x). This is because with larger applications (ones with more clusters), the proposed approach has greater scope to improve the inference lifetime by intelligently mapping the synaptic weights in all the clusters.

\begin{figure}[h!]
	\begin{center}
		\vspace{-15pt}
		\includegraphics[width=0.99\columnwidth]{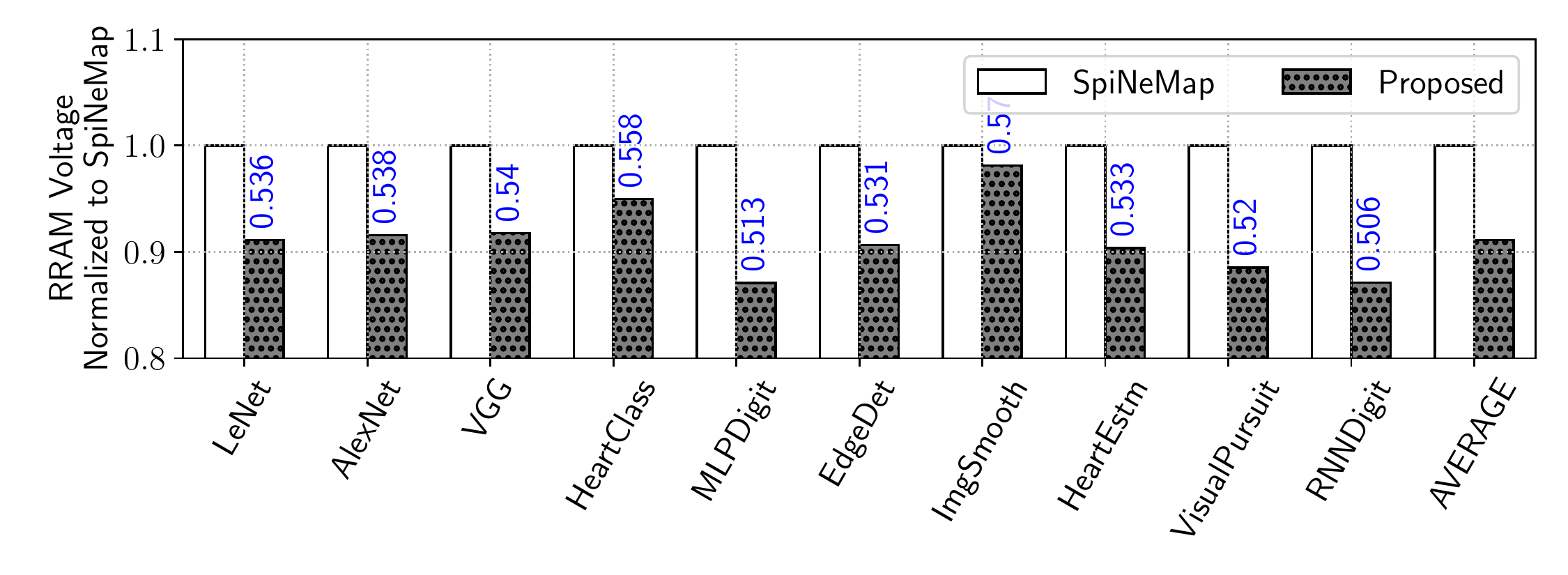}
		\vspace{-25pt}
		\caption{Average RRAM voltage normalized to SpiNeMap.}
		\vspace{-15pt}
		\label{fig:current_value}
	\end{center}
\end{figure}
To give further insight, Figure~\ref{fig:current_value} plots the average voltage on the RRAM cells within a crossbar when the clusters of each evaluated application are mapped to them. For reference, we have reported the absolute voltage in V for the proposed approach. We observe that the average voltage on the RRAM cells in the proposed approach is 9\% lower than \sm{}. This is because the proposed approach uses the top right corners of a crossbar (see Figure~\ref{fig:parasitics}) to place the synaptic weights. This is where the parasitic voltage drops are higher, resulting in a lower voltage across the RRAM cells.

\subsection{Spike Delay}
\begin{figure}[h!]
	\begin{center}
		\vspace{-15pt}
		\includegraphics[width=0.99\columnwidth]{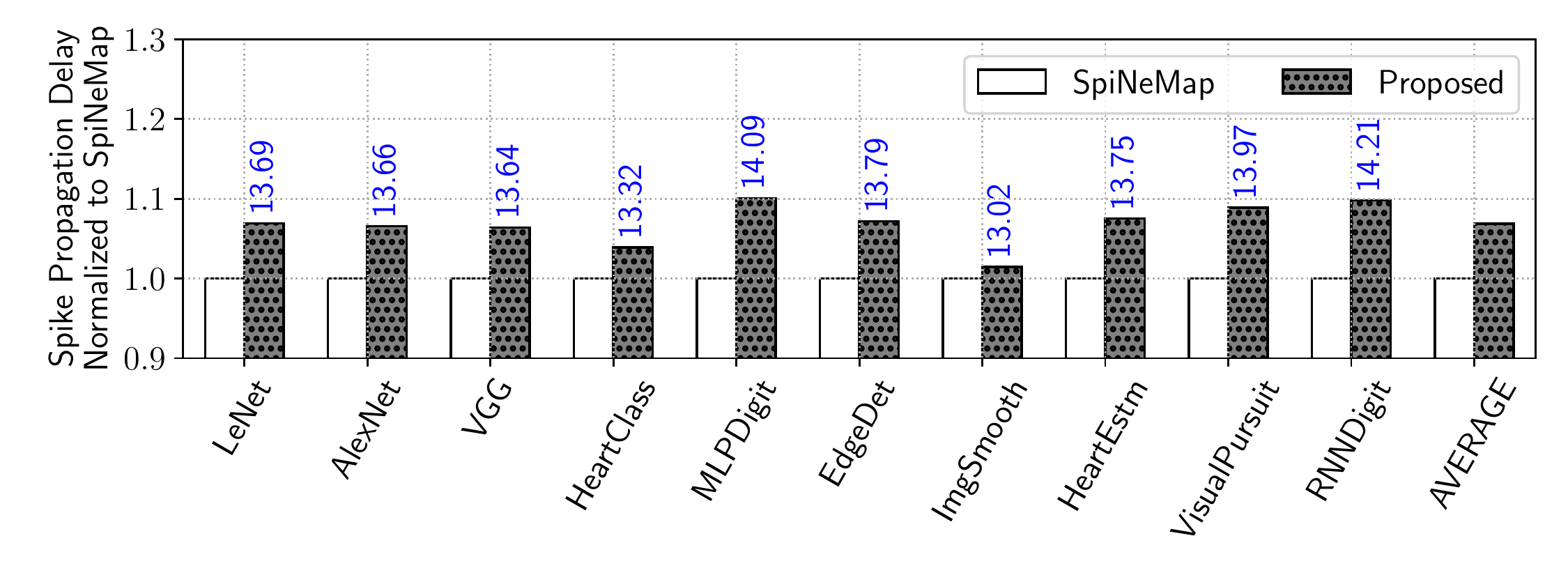}
		\vspace{-25pt}
		\caption{Crossbar spike propagation delay normalized to SpiNeMap.}
		\vspace{-15pt}
		\label{fig:delay_value}
	\end{center}
\end{figure}

Unfortunately, the RRAM cells at the top right corner of each crossbar introduce longer spike propagation delay than those at the bottom left corner. To estimate the average increase in spike delay, Figure~\ref{fig:delay_value} plots the spike propagation delay through the crossbar for each application, normalized to \sm{}. For reference, we have reported the absolute delay in ms using the proposed approach. We observe that the spike propagation delay using the proposed approach is only an average 6\% higher than \sm{}.

\subsection{Inference Lifetime with Limited Hardware Resources}
\begin{figure}[h!]
	\begin{center}
		\includegraphics[width=0.99\columnwidth]{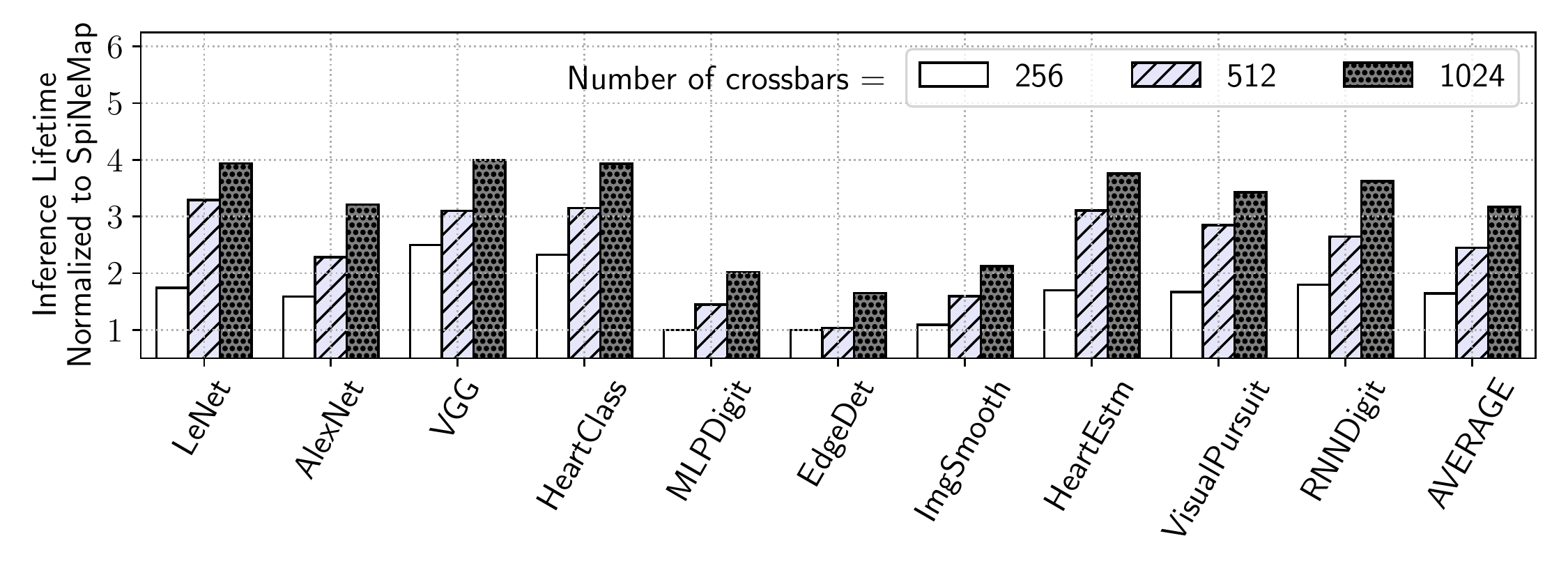}
		\vspace{-25pt}
		\caption{Inference lifetime for three different hardware configurations.}
		\vspace{-15pt}
		\label{fig:il_limited}
	\end{center}
\end{figure}

Figure~\ref{fig:il_limited} plots the inference lifetime obtained using the proposed approach normalized to \sm{} as we increase the hardware size from 256 crossbars to 1024 crossbars. We make the following two observations.

First, the inference lifetime of the proposed approach increases with an increase in the size of the hardware. With 256, 512, and 1024 crossbars in the hardware, the inference lifetime of the proposed approach is higher than \sm{} by an average of 1.64x, 2.45x, and 3.16x. With fewer crossbars in the hardware, the number of clusters mapped to each hardware increases, increasing the crossbar utilization. Therefore, the proposed approach has limited scope to reorganize the synapses onto the RRAM cells, resulting in lower improvement than the case where there are more crossbars in the hardware. Second, the improvement of inference lifetime in smaller applications like MLPDigit, EdgeDet, and ImgSmooth is not significant compared to larger applications like LeNet, AlexNet, and VGG. From these results, we conclude that the proposed approach has a greater opportunity to increase the inference lifetime for crossbars with lower utilization.

\subsection{Exploration Time}
Table~\ref{tab:exploration_time} reports the exploration time of the proposed Hill-Climbing-based mapping exploration for each of the evaluated applications. Column 2 reports the number of clusters of these applications generated using \sm{}~\cite{spinemap}. For these clusters, Columns 3, 4, and 5 report the exploration time for three hardware configurations -- 256 crossbars, 512 crossbars, and 1024 crossbars, respectively. We make the following two observations. First, the exploration time increases with increase in the number of crossbars due to the increase in the size of the search space. Second, for applications such as ImgSmooth and RNNDigit, there is no significant increase in the exploration time because the number of clusters for these applications is less than the number of crossbars in the hardware. Therefore, the application mapping time is essentially the time in solving the BNLP problem.

\vspace{-10pt}
\begin{table}[h!]
	\renewcommand{\arraystretch}{0.8}
	\setlength{\tabcolsep}{2pt}
	\caption{Exploration time of the proposed approach.}
	\label{tab:exploration_time}
	\vspace{-5pt}
	\centering
	\begin{threeparttable}
	{\fontsize{6}{10}\selectfont
		\begin{tabular}{cc|ccc}
			\hline
			\textbf{Applications} &
			\textbf{Clusters} &
			\textbf{256 crossbars} & \textbf{512 crossbars} & \textbf{1024 crossbars}\\
			\hline
			LeNet & 2,066 & 2,189 sec & 2,702 sec & 3, 183 sec\\
			AlexNet & 30,105 & 40,667 sec & 61,434 sec & 101,924 sec\\
			VGG & 95,452 & 70,180 sec & 144,090 sec & 389, 644 sec \\
			HeartClass & 7,871 & 5,111 sec & 8,110 sec & 12, 145 sec\\
			MLPDigit & 520 & 594 sec & 784 sec & 1,010 sec\\
			EdgeDet & 437 & 405 sec &  612 sec & 824 sec\\
			ImgSmooth & 41 & 52 sec & 52 sec & 52 sec\\
 			HeartEstm & 325 & 321 sec & 486 sec & 611 sec\\
 			VisualPursuit & 1,001 & 1,138 sec & 1,520 sec & 1,921 sec\\
 			RNNDigit & 90 & 114 sec & 114 sec & 114 sec\\
			\hline
	\end{tabular}}
	\end{threeparttable}
\end{table}
\vspace{-10pt}

\subsection{Technology Scaling}
\begin{figure}[h!]
	\begin{center}
		\vspace{-15pt}
		\includegraphics[width=0.99\columnwidth]{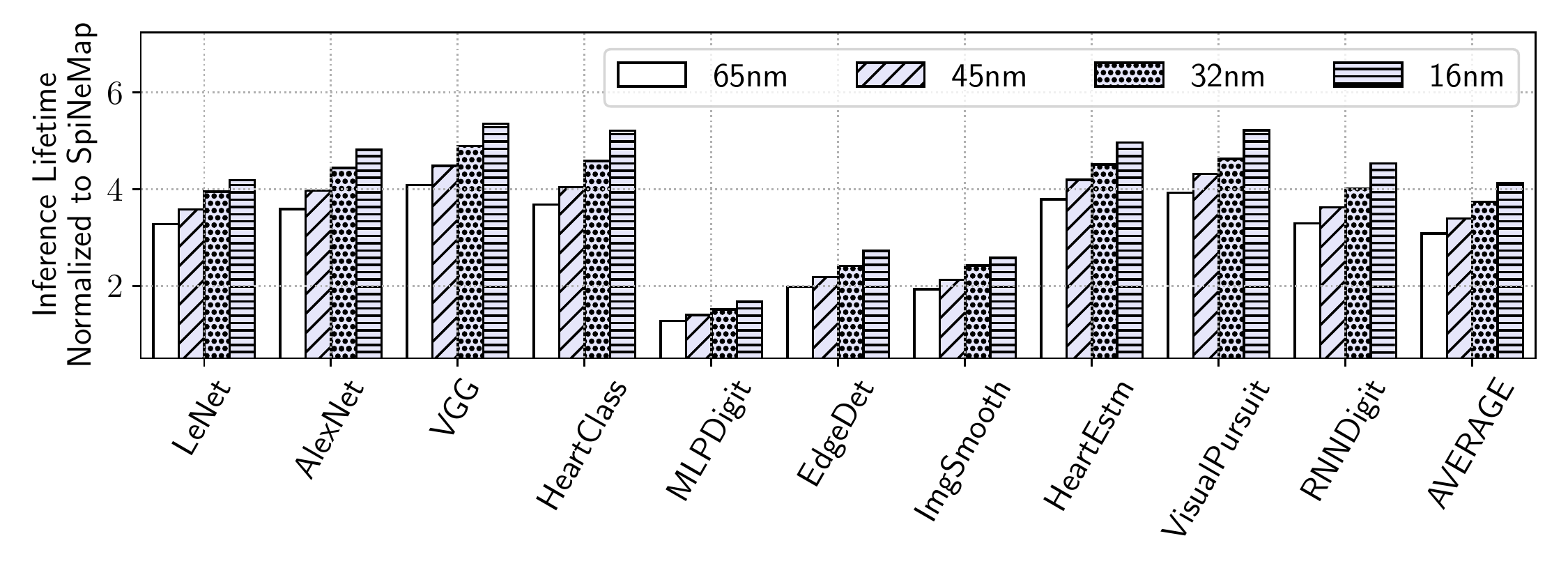}
		\vspace{-25pt}
		\caption{Impact of technology scaling on inference lifetime.}
		\vspace{-15pt}
		\label{fig:technology}
	\end{center}
\end{figure}

Figure~\ref{fig:technology} plots the inference lifetime of the proposed approach normalized to \sm{} for four technology nodes -- 65nm, 45nm (default), 32nm, and 16nm. We observe that the improvement of inference lifetime over \sm{} increases as the technology scales down, even though the absolute inference lifetime is lower at scaled nodes. This is because with technology scaling, the endurance variation within each crossbar becomes more significant. Therefore, the proposed approach, which incorporates such variation in the cluster mapping and synapse placement process leads to higher inference lifetime compared to \sm{}.

%% file: sections/conclusions.tex
We present a novel Binary Non-Linear Programming (BNLP) formulation of the inference lifetime of machine learning workloads when mapped on to the RRAM cells of a neuromorphic system. 
Using such formulation, we show that the parasitic IR drops in the system create a significant difference in read endurance of the RRAM cells. We incorporate the BNLP formulation and endurance variation inside a Hill-Climbing-based mapping exploration to find an optimum mapping of the clusters of an inference model to the crossbars of a hardware, improving its inference lifetime. Our formulation ensures that critical synapses (those that propagate more spikes) are never mapped on to the weaker cells (ones that have lower endurance). We evaluate our approach with 10 machine learning applications on a cycle-accurate simulator of state-of-the-art neuromorphic hardware. Our results demonstrate an average 3.4x improvement in inference lifetime with only 6\% increase in spike propagation delay.